\documentclass[sigconf]{acmart}
\usepackage{booktabs}
\usepackage{graphicx}
\usepackage{stfloats}
\usepackage{comment}
\usepackage{multirow}
\usepackage{color}
\usepackage{capt-of}
\usepackage{bm}
\usepackage{bbm}
\usepackage{verbatim}
\newcommand{\blank}[1]{\hspace*{#1}\linebreak[0]}
\usepackage[ruled,linesnumbered]{algorithm2e}
\usepackage{float}
\usepackage{colortbl}
\usepackage{arydshln}
\usepackage{multirow}
\usepackage{multicol}

\DeclareMathOperator*{\argmax}{arg\,max}

\AtBeginDocument{%
  \providecommand\BibTeX{{%
    \normalfont B\kern-0.5em{\scshape i\kern-0.25em b}\kern-0.8em\TeX}}}

\setcopyright{rightsretained}
\copyrightyear{2021}
\acmYear{2021}
\acmDOI{xxx}

\acmConference[WWW '21]{Proceedings of the Web Conference 2021}{April 19--23, 2021}{Ljubljana, Slovenia}
\acmBooktitle{Proceedings of the Web Conference 2021 (WWW '21), April 19--23, 2021,
Ljubljana, Slovenia}
\acmPrice{}
\acmISBN{978-x-xxxx-xxxx-x/YY/MM}

\begin{document}

\title{A Hybrid Bandit Model with Visual Priors for Creative Ranking in Display Advertising}



\author{Shiyao Wang}
\affiliation{%
  \institution{Alibaba Group}
  \city{Beijing}
  \country{China}}
\email{shiyao.wsy@alibaba-inc.com}

\author{Qi Liu$^*$}
\affiliation{%
  \institution{University of Science and Technology of China}
  \city{Hefei}
  \country{China}}
\email{qiliu67@mail.ustc.edu.cn}

\author{Tiezheng Ge}
\affiliation{%
  \institution{Alibaba Group}
  \city{Beijing}
  \country{China}}
\email{tiezheng.gtz@alibaba-inc.com}

\author{Defu Lian}
\affiliation{%
  \institution{University of Science and Technology of China}
  \city{Hefei}
  \country{China}}
\email{liandefu@ustc.edu.cn}

\author{Zhiqiang Zhang}
\affiliation{%
  \institution{Alibaba Group}
  \city{Beijing}
  \country{China}}
\email{zhang.zhiqiang@alibaba-inc.com}



\begin{abstract}
Creative plays a great important role in e-commerce for exhibiting products. Sellers usually create multiple creatives for comprehensive demonstrations, thus it is crucial to display the most appealing design to maximize the Click-Through Rate~(CTR). For this purpose, modern recommender systems dynamically rank creatives when a product is proposed for a user. However, this task suffers more cold-start problem than conventional products recommendation since the user-click data is more scarce and creatives potentially change more frequently.
In this paper, we propose a hybrid bandit model with visual priors which first makes predictions with a visual evaluation, and then naturally evolves to focus on the specialities through the hybrid bandit model.
Our contributions are three-fold:
1) We present a visual-aware ranking model (called VAM) that incorporates a list-wise ranking loss for ordering the creatives according to the visual appearance.
2) Regarding visual evaluation as a prior, the hybrid bandit model (called HBM) is proposed to evolve consistently to make better posteriori estimations by taking more observations into consideration for online scenarios.
3) A first large-scale creative dataset, \textit{CreativeRanking}\textsuperscript{\ref {github}}, is constructed, which contains over 1.7M creatives of 500k products as well as their real impression and click data.
Extensive experiments have also been conducted on both our dataset and public Mushroom dataset, demonstrating the effectiveness of the proposed method.
\end{abstract}



\keywords{Hybrid Bandit Model, Visual Priors, Creative Ranking}


\maketitle

\section{Introduction}
\begin{figure}[t]
\setlength{\belowcaptionskip}{0cm}
\setlength{\abovecaptionskip}{0.4cm} 
\begin{center}
\includegraphics[width=1\linewidth]{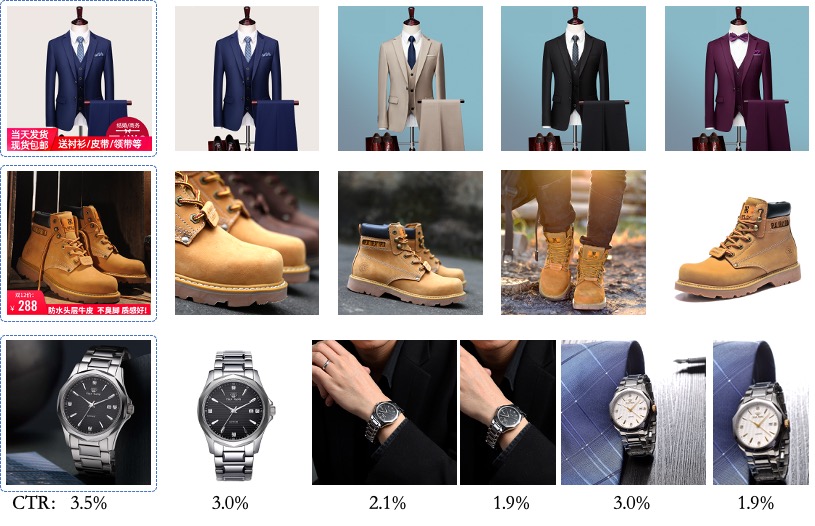}
\end{center}
\caption{Some examples of ad creatives. Each row presents creatives that display the product in multiple ways. The corresponding CTRs at the bottom row indicate the large CTR gap among creatives. }
\label{fig1}
\end{figure}

Online display advertising is a rapid growing business and has become an important source of revenue for Internet service providers. The advertisements are delivered to customers through various online channels, e.g. e-commerce platform. Image ads are the most widely used format since they are more compact, intuitive and comprehensible \cite{chen2016deep}. 
In Figure \ref{fig1}, each row composes several ad images that describe the same product for comprehensive demonstrations. These images are called creatives.
Although the creatives represent the same product, they may have largely different CTRs due to the visual appearance. Thus it is crucial to display the most appealing design to attract the potentially interested customers and maximize the Click-Through Rate(CTR).

In order to explore the most appealing creative, all of the candidates should be displayed to customers. Meanwhile, to ensure the overall performance of advertising, we prefer to display the creative that has the highest predicted CTR so far. \textit{This procedure can be modeled as a typical multi-armed bandit problem (MAB)}. It not only focuses on maximizing cumulative rewards (clicks) but also balance the exploration-exploitation(E\&E) trade-off within a limited exploration resource so that CTR are considered. Epsilon-greedy \cite{sutton2018reinforcement}, Thompson sampling \cite{russo2014learning} and Upper Confidence Bounds (UCB) approaches \cite{auer2002finite} are widely used strategies to deal with the bandit problem. \textbf{However, creatives potentially change more frequently than products, and most of them cannot have sufficient impression opportunities to get reliable CTRs throughout their lifetime.} So the conventional bandit models may suffer from cold-start problem in the initial random exploration period, hurting the online performance extremely. One potential solution to this problem is incorporating visual prior knowledge to facilitate a better exploration. 
\cite{azimi2012impact,ChengZAMZZN12,mo2015image,chen2016deep} consider the visual features extracted by deep convolutional networks and make deterministic selections for product recommendation. These deep models are in a heavy computation and cannot be flexibly updated online. Besides, the deterministic and greedy strategy may result in suboptimal solution due to the lack of exploration. Consequently, how to combine both the expressive visual representations and flexible bandit model remains a challenging problem.

In this paper, we propose an elegant method which incorporates visual prior knowledge into bandit model for facilitating a better exploration. It is based on a framework called NeuralLinear \cite{riquelme2018deep}. They consider approximate bayesian neural networks in a Thompson sampling framework to utilize both the learning ability of neural networks and the posterior sampling method. By adopting this general framework,
we first present a novel convolutional network with a list-wise ranking loss function to select the most attractive creative. The ranking loss concentrates on capturing the visual patterns related to attractiveness, and the learned representations are treated as contextual information for the bandit model. Second, in terms of the bandit model, we make two major improvements: 1) Instead of randomly setting a prior hyperparameter to candidate arms, we use the weights of neural network to initialize the bandit parameters that further enhance the performance in the cold-starting phase. 2) To fit the industrial-scale data, we extend the linear regression model of NeuralLinear to a hybrid model which adopts two individual parameters, i.e. product-wise ones and creative-specific ones. The two components are adaptively combined during the exploring period. 
Last but not the least, because the creative ranking is a novel problem, it lacks real-world data for further study and comparison. To this end, we contribute a large-scale creative dataset\footnote{The Data and code are publicly available at https://github.com/alimama-creative/A\_Hybrid\_Bandit\_Model\_with\_Visual\_Priors\_for\_Creative\_Ranking.git\label{github}} from Alibaba display advertising platform that comprises more than 500k products and 1.7M ad creatives. 


In summary, the contributions of this paper include:

- We present a visual-aware ranking model (called VAM) that is capable of evaluating new creatives according to the visual appearance.

- Regarding the learned visual predictions as a prior, the improved hybrid bandit model (called HBM) is proposed to make better posteriori estimations by taking more observations into consideration.

- We construct a novel large-scale creative dataset named \textit{CreativeRanking}\textsuperscript{\ref {github}}.  Extensive experiments have been conducted on both our dataset and public Mushroom dataset, demonstrating the effectiveness of the proposed method.

\section{Preliminaries and Related Work}
\subsection{Preliminaries}
\textbf{Problem Statement} The problem statement is as follows. Given a product, the goal is to determine which creative is the most attractive one and should be displayed. Meanwhile, we need to estimate the uncertainty of the predictions so as to maximize cumulative reward in a long run.

In the online advertising system, when an ad is shown to a user by displaying a candidate creative, this scenario is counted as an impression. 
Suppose there are $N$ products, denoted as $\{I^1, I^2, \cdots, I^n, \\\cdots, I^N\}$, and each product $I^n$ composes a group of creatives, indicated as $\{C_1^n, C_2^n, \cdots, C_m^n, \cdots, C_M^n\}$. For product $I^n$, the objective is to find the creative that subjects to:
\begin{equation}\label{equ1}
C^n = \argmax_{ c\in \{C_1^n, C_2^n, \cdots, C_M^n\}} CTR(c) \\
\end{equation}
where $CTR(\cdot)$ denotes the CTR for a given creative. An empirical way to produce CTR is accumulating the current clicks and impressions, and produce the click ratio as: 
\begin{equation}\label{equ2}
\hat{CTR}(C_m^n) = \frac{click(C_m^n)}{impression(C_m^n)}\\
\end{equation}
where $click(\cdot)$ and $impression(\cdot)$ indicate the click and impression number of the creative $C_m^n$. But it may suffer from insufficient impressions, especially for the cold-start creatives. Another way is to learn a prediction function $\mathcal{N}(\cdot)$ from all the historical data by considering the contextual information (i.e. the image content) as: 
\begin{equation}\label{equ3}
\hat{CTR}(C_m^n) = \mathcal{N}(C_m^n)\\
\end{equation}
where $\mathcal{N}(\cdot)$ takes the image content of creative $C_m^n$ as input, and learns from the historical data. The collected sequential data can be represented as 
\begin{equation}\label{data}
\mathcal{D} = \{(C_1, y_1), \cdots, (C_t, y_t), \cdots, (C_{|\mathcal{D}|}, y_{|\mathcal{D}|})\}
\end{equation}
 where $y_t \in \{0, 1\}$ is the label denotes whether a click is received. We take both the statistical data and content information into consideration. Subsection \ref{visual} reviews some product recommendation methods that take visual content as auxiliary information, and subsection \ref{bandit} introduces typical bandit models to estimate uncertainty. Both of above methods will be our strong baselines.

\subsection{Visual-aware Recommendation Methods}\label{visual}
CTR prediction of image ads is a core task of online display advertising systems. Due to the recent advances in computer vision, visual features are employed to further enhance the recommendation models \cite{azimi2012impact,ChengZAMZZN12,mo2015image,chen2016deep,wang2018telepath,ge2018image,yu2018aesthetic,capelo2019combining,liu2020category}. \cite{azimi2012impact,ChengZAMZZN12} quantitatively study the relationship between handcrafted visual features and creative online performance. Different from fixed handcrafted features, \cite{ge2018image,yu2018aesthetic,capelo2019combining} apply ``off-the-shelf'' visual features extracted by deep convolutional neural network\cite{simonyan2014very}. \cite{mo2015image,chen2016deep,wang2018telepath} extend these methods by training the CNNs in an end-to-end manner. \cite{liu2020category} integrate the category information on top of the CNN embedding to help visual modeling.
The above works focus on improving the \textbf{product ranking} by considering visual information while neglecting the great potential of \textbf{creative ranking}. There is a few work so far to address this topic. idealo.de (portal of the German e-commerce market) adopts an aesthetic model\cite{talebi2018nima} to select the most attractive image for each recommended hotel. They believe that photos can be just as important for bookings as reviews. PEAC \cite{zhao2019you} resembles our method the most and they aim to rank ad creatives based on the visual content. But it is an offline evaluation model that cannot flexibly update the ranking strategy when receiving online observations. Besides, all above methods do \textit{not} model the uncertainty which may lack of exploration ability.

\subsection{Multi-armed Bandit Methods}\label{bandit}
Multi-armed bandits (MAB) problem is a typical sequential decision making process that is also treated as an online decision making problems \cite{yang2020hierarchical}. A wide range of real world applications can be modeled as MAB problems, such as online recommendation system \cite{glowacka2019bandit}, online advertising \cite{schwartz2017customer} and information retrieval \cite{glowacka2017bandit}. Epsilon-greedy \cite{sutton2018reinforcement}, Thompson sampling \cite{russo2014learning} and UCB \cite{auer2002finite} are classic context-free algorithms. They use reward/cost from the environment to update their E\&E policy without contextual information. It is difficult for model to quickly adjust to new creatives (arms) since the web content undergoes frequent changes. \cite{li2010contextual,agrawal2013thompson,riquelme2018deep} extend these context-free methods by considering side information like user/content representations. They assume that the expected payoff of an arm is linear in its features. The main problem linear algorithms face is their lack of representational power, which they complement with accurate uncertainty estimates. A natural attempt at getting the best of both representation learning ability and accurate uncertainty estimation consists in performing a linear payoffs on top of a neural network. NeuralLinear \cite{riquelme2018deep} present a Thompson sampling based framework that simultaneously learn a data representation through neural networks and quantify the uncertainty over Bayesian linear regression. Inspired by this framework, we further improve both the neural network and bandit method that benefit our creative ranking problem.

\begin{figure*}[t]
\vspace{-0.1cm}
\setlength{\abovecaptionskip}{0.1cm} 
\setlength{\belowcaptionskip}{-0.3cm} 
\centering
\includegraphics[width=1.0\linewidth]{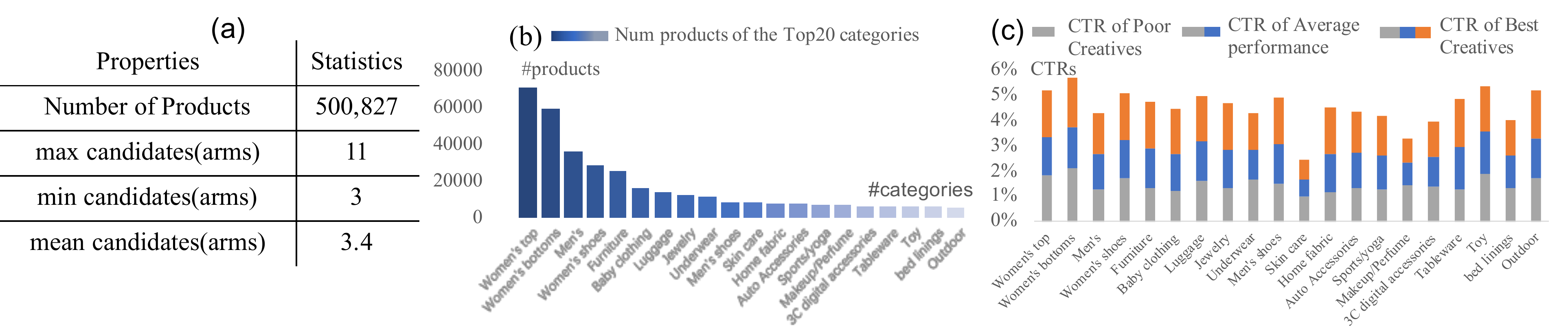}
\caption{Statistical Analysis of the \textit{Creative Ranking} dataset. (a) summarizes some basic information, while (b) shows the number of products in terms of product categories. (c) conducts CTR analysis by comparing poor and good creatives.}
\label{statis}
\end{figure*}

\section{Dataset Construction}
In order to promote further research and comparison on creative ranking, we contribute a large-scale creative dataset to the research community. It composes creative images and sequential impression data which can be used for evaluating both visual predictions and E\&E strategies.
In this section, we first describe how the creatives and online feedbacks are collected in subsection \ref{data_collection}. Then we provide a statistical analysis of the dataset in subsection \ref{statistics}. 

\subsection{Data Collection}\label{data_collection}

We collect a large and diverse set of creatives from Alibaba display advertising platform during July 1, 2020 to August 1, 2020. The total number of impression is approximately 215 million. There are 500,827 products with 1,707,733 ad creatives. We make this dataset publicly available for further research and comparison. The creative and online feedback collection is subject to the following constraints:

\textbf{Randomized logging policy.} The online system adopts randomized logging policy so that the creatives are randomly drawn to collect an unbiased dataset. Bandit algorithms learn policies through interaction data. Training or evaluation on offline data may suffer from exposure bias called "off-policy evaluation problem" \cite{precup2000eligibility}. In \cite{li2010contextual}, they demonstrate that if logging policy chooses each arm uniformly at random, the estimation of bandit algorithms is unbiased.  Thus, for each impression of product $I^n$, the policy will randomly choose a candidate creative, and gather their clicks. 

\begin{figure}[h]
\vspace{-0.3cm}
\setlength{\abovecaptionskip}{-0.0cm} 
\setlength{\belowcaptionskip}{-0.3cm} 
\begin{center}
\includegraphics[width=0.9\linewidth]{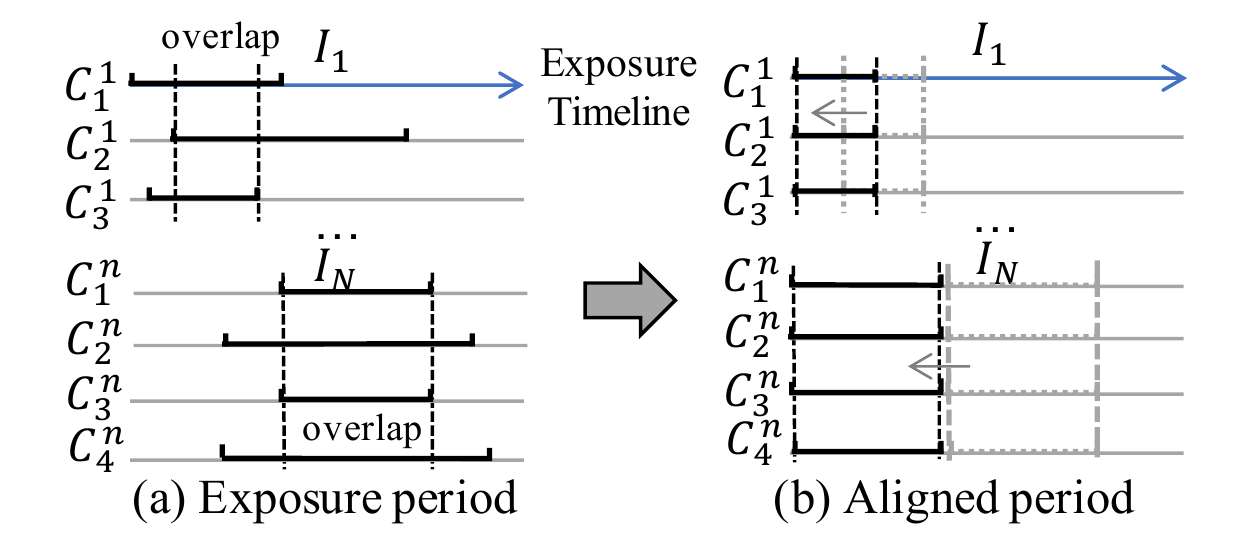}
\end{center}
\caption{Aligned creative lifetime. }
\label{lifetime}
\end{figure}

\textbf{Aligned creative lifetime. } Due to the complexity of online environment, the CTRs vary for different time periods, even for the same creative. Creatives will be newly designed or deleted, which will result to inconsistent exposure time (as Figure \ref{lifetime}(a)). In order to avoid the noise brought by the different time intervals, we only collect the overlap period among the candidate creatives (see Figure \ref{lifetime}(b)). Besides, the overlap should be within 5 to 14 days, which covers the creative lifetime from the cold-starting to relative stable stage. All the filtered creatives are gathered to build the sequential data.

\textbf{Train/Validation/Test split.} We randomly split the 500,827 products into 300,242 training, 100,240 validation and 100,345 test samples, with 1,026,378/340,449/340,906 creatives respectively. We treat each product as a sample, and aim to select the best creative among candidates. The proposed VAM is learned from the training set, while the bandit model HBM is deployed on the validation/test data. This setting is used to prove the effectiveness of visual predictions on the unseen products/creatives, and whether the policy can make a better posterior estimations by using online observations.

\subsection{Statistical Analysis}\label{statistics}
The proposed dataset is collected from ad interaction logs across 32 days. Figure \ref{statis}(a) gives a summary of our CreativeRanking dataset. It consists of 500,827 products, covering 124 categories. The min and max candidate creatives for a product is 3 and 11, while average number is 3.4. In fact, the number of candidates in the real-world scenarios far exceeds 3.4, but the offline dataset is constrained by conditions introduced by subsection \ref{data_collection}.
Figure \ref{statis}(b) shows the number of products for top 20 categories, namely Women's tops, Women's bottoms, Men's, Women's shoes, Furniture, and so on. In Figure \ref{statis}, we make further analysis about creatives for these categories. Suppose we know the CTR for each creative, we select the poorest and best creatives for each product, and accumulate their overall performance, which is visualized as grey and (grep+blue+orange) bins. We find that the CTR of a product can be extremely lifted by selecting a good creative. Specifically, a good creative is capable of lifting CTR by 148\% $\sim$ 285\% compared to the poorest candidates, while it turns to 41.5\% $\sim$ 72.5\% compared to averaged performance of all candidates (grep+blue bins). 

By proposing this \textit{CreativeRanking} dataset, we would like to draw more attention to this topic which benefits both the research community and website's user experience.

\begin{figure*}[ht]
\vspace{-0.1cm}
\setlength{\abovecaptionskip}{0.3cm} 
\setlength{\belowcaptionskip}{-0.1cm}
\centering
\includegraphics[width=1\linewidth]{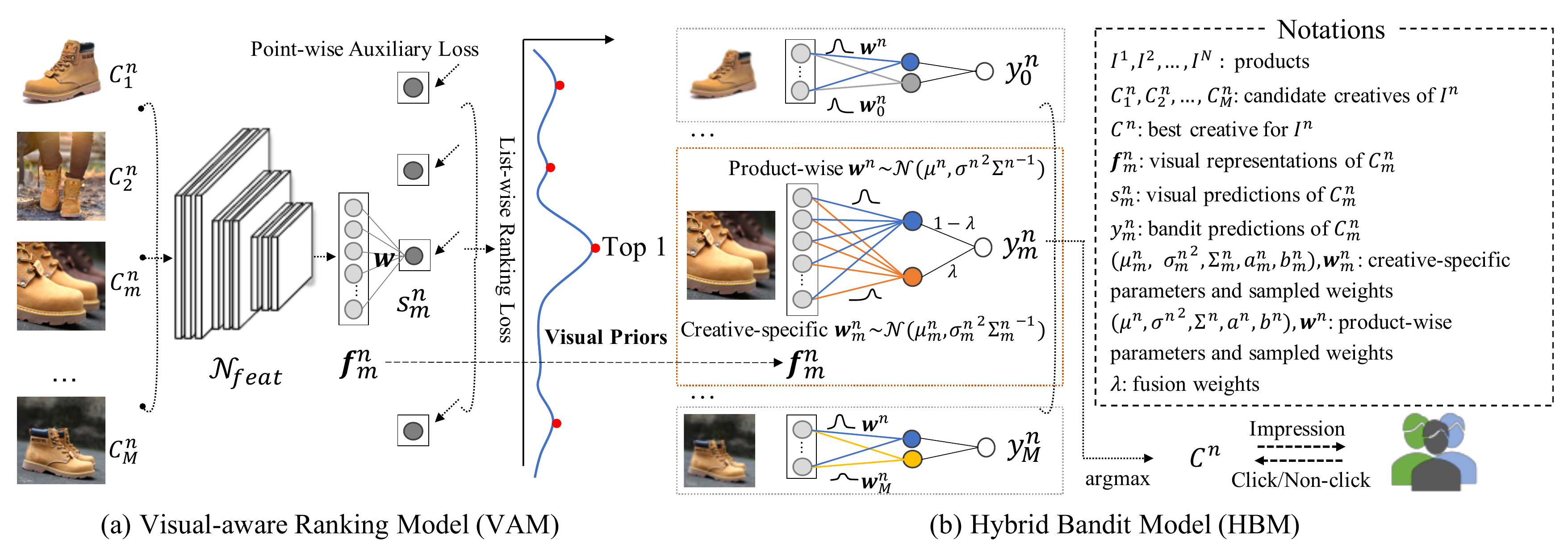}
\caption{(Better viewed in color) The overall framework of the proposed Hybrid Bandit Model with Visual Priors. It receives several candidate creatives (shown in one column on the left) and try to find the most attractive one through both Visual-aware Ranking Model (VAM) and Hybrid Bandit Model (HBM). (a) VAM is to develop a CNN model that can evaluate creatives base on their visual content. (b) According to the visual priors, HBM aims to estimate the posterior and correct the ranking strategy.}
\label{fig2}
\end{figure*}
\section{Method}

\subsection{Overview}
We briefly overview the entire pipeline. Main notations used in this paper are summarized in the right panel of Figure \ref{fig2}. 
First, as shown in Figure \ref{fig2}(a), feature extraction network $\mathcal{N}_{feat}$ will simultaneously receive the creatives of the $n-$th product as input, and produce the $d$-dimensional intermediate features $\{\bm{f}_1^n, \bm{f}_2^n, \cdots, \bm{f}_m^n, \cdots, \bm{f}_M^n\}$. Then, a fully connected layer are employed to calculate the scores for them, indicated as $\{s_1^n, s_2^n,\cdots, s_m^n, \cdots, s_M^n\}$.

Second, the \textit{list-wise ranking loss and auxiliary regression loss} are introduced to guide the learning procedure. Such a multi-objective optimization helps the model not only focus on creative ranking, but also take into account the numerical range of CTR that is benefit for the following bandit model. In addition, due to the data noise that is a common problem in a real-world application, we provide several practical solutions to mitigate casual and malicious noise. Details are described in Subsection \ref{VAM}.

After the above steps, the model can evaluate the creative quality directly by its visual content, even a newly uploaded one without any history information. Then we propose a hybrid bandit model that incorporates learned $\bm{f}_m^n$ as contextual information,  and update the policy by interacting with online observations. As in Figure \ref{fig2}(b), 
the hybrid model combines both product-wise and creative-specific predictions which is more flexible for complex industrial data. The elaborated formulations are in Subsection \ref{HBM}.

\subsection{VAM: Visual-aware Ranking Model}
\label{VAM}
Given a product $I^n$, we use feature extraction network $\mathcal{N}_{feat}$ to extract high-level visual representations of creatives. And a linear layer is adopted to produce the attractiveness scores for $m$-th creative of $n$-th product:
\begin{equation}\label{equ4}
\bm{f}_m^n = \mathcal{N}_{feat}(C_m^n)\\
\end{equation} 
\begin{equation}\label{equ5}
s_m^n = \bm{f}_m^{nT}\bm{w}\\
\end{equation}
where $\bm{w}$ are learnable parameters of the linear layer.

\textbf{List-wise Ranking Loss}. To learn the relative order of creatives, we need to map a list of predicted scores and ground-truth CTRs to a permutation probability distribution, respectively, and then take a metric between these distributions as a loss function. The mapping strategy and evaluation metric should guarantee that the candidates with higher scores would be ranked higher. \cite{cao2007learning} proposed permutation probability and top $k$ probability definitions. Inspired by this work, we simplify the probability of a creative being ranked on the top 1 position as 
\begin{equation}\label{equ6}
p_m^n = \frac{exp(s_m^n)}{\sum_{i=1}^M{exp(s_i^n)}}\\
\end{equation}
where $exp(\cdot)$ is an exponential function. The exponential function based top-1 probability is both scale invariant and translation invariant. And the corresponding labels are
\begin{equation}\label{equ7}
{y}_{rank}(C_{m}^{n}) = \frac{exp(CTR(C_m^n), T)}{\sum_{i=1}^M{exp(CTR(C_i^n), T)}}\\
\end{equation}
where $exp(\cdot, T)$ is exponential function with temperature $T$. Since the $CTR(C_m^n)$ is about a few percent, we use $T$ to adjust the scale of the value so that make the probability of top1 sample close to 1. With Cross Entropy as metric, the loss for product $I^n$ becomes
\begin{equation}\label{equ8}
\mathcal{L}_{rank}^{n} = -\sum_m{{y}_{rank}(C_{m}^{n})log(p_m^n)}\\
\end{equation}

Through such objective function, the model focuses on comparing the creatives within the same product. 
We concentrate on the top-1 probability since it is consistent with real scenarios which will display only one creative for each impression. Besides, the end-to-end training manner greatly utilizes the learning ability of deep CNNs and boosts the visual prior knowledge extraction.

\textbf{Point-wise auxiliary regression Loss}. In addition to the list-wise ranking loss, we expect that the point-wise regression enforce the model to produce more accurate predictions. Actually, the ranking loss function only requires the order of outputs, leaving the numerical scale of the outputs unconstrained. Since the learned representations will be adopted as prior knowledge for the bandit model in Subsection \ref{HBM}, making the outputs close to the real CTRs will significantly stabilize the bandit learning procedure. Thus we add the point-wise regression as a regularizer. The formulation is 
\begin{equation}\label{equ9}
\mathcal{L}_{reg}^{n} =\sum_m{||CTR(C_m^n)- s_m^n||_2}\\
\end{equation}
where $||\cdot||$ denotes $L_2$ norm. Finally, we add up both the ranking loss and the auxiliary loss to form the final loss:
\begin{equation}\label{equ10}
\mathcal{L}^{n} = \mathcal{L}_{rank}^{n}+ \gamma \mathcal{L}_{reg}^{n}\\
\end{equation}
where $\gamma$ is 0.5 in our experiments. 

\textbf{Noise Mitigation}. In both list-wise ranking and point-wise regression in Equation \ref{equ7} and \ref {equ9}, ${CTR}(C_m^n)$ can be estimated by Equation \ref{equ2}.
But in real-world dataset, some creatives have not sufficient impression opportunities, and the estimation may suffer from huge variance. For example, a creative only get one impression, and a click is accidentally recorded from this impression, the $\hat{CTR}$ will be set to 1, which is inevitably unreliable. To mitigate the problem, we provide two practical solutions, namely label smoothing and weighted sampling. 

\textit{Label smoothing} is an empirical Bayes method that is utilized to smoothen the CTR estimation \cite{wang2011click}. 
Suppose the clicks are from a binomial distribution and the CTR follows a prior distribution as 
\begin{equation}\label{equ11}
\begin{split}
clicks(C_m^n) &\sim Binomial(Impression(C_m^n), CTR(C_m^n))\\
CTR(C_m^n) &\sim Beta(\alpha, \beta)\\
\end{split}
\end{equation}
where $Beta(\alpha, \beta)$ can be regarded as the prior distribution of CTRs. After observing more clicks, the conjugacy between Binomial and Beta allows us to obtain the posterior distribution and  the smoothed $\hat{CTR}$ as 
\begin{equation}\label{equ12}
\hat{CTR}(C_m^n) = \frac{click(C_m^n)+\alpha}{impression(C_m^n)+\alpha+\beta}\\
\end{equation}
where $\alpha$ and $\beta$ can be yielded by using maximum likelihood estimate through all the historical data\cite{wang2011click}. Compared to the original way, the smoothed $\hat{CTR}$ has smaller variance and benefits the training.

\textit{Weighted sampling} is a sampling strategy for training process. Instead of treating each product equally, we pay more attention to the products whose impressions are adequate and the CTRs are more reliable. The sampling weights can be produced by 
\begin{equation}\label{equ_ws}
p^n = g(impression(I^n))\\
\end{equation}
where $g(\cdot)$ is set to the logarithm of the impressions and $p^n$ denotes the sampling weight of product $I^n$. 

All above modules are integrated in a unified framework and the visual-aware ranking model focuses on learning the general visual patterns about display performance. And then the informative representations are applied as prior knowledge for the bandit algorithm.

\subsection{HBM: Hybrid Bandit Model}
\label{HBM}
In this section, we provide an elegant and efficient strategy that tackles the E\&E dilemma by utilizing the visual priors and updating the posterior via the hybrid bandit model. 
Based on NeuralLinear framework \cite{riquelme2018deep}, we build a Bayesian linear regression on the extracted visual representation. Assume the online feedback data is generated as follows:
\begin{equation}\label{equ13}
\bm{y} = \bm{f}^{T}\tilde{\bm{w}} + \epsilon\\
\end{equation}
where $\bm{y}$ represent clicked/non-clicked data and $\bm{f}$ is the extracted visual representations by VAM. Different from the deterministic weights $\bm{w}$ in Equation \ref{equ5}, we need to learn a weight distribution $\tilde{\bm{w}}$ with the uncertainty that benefits the E\&E decision making. $\epsilon$ are independent and identically normally distributed random variables:
\begin{equation}\label{epsilon}
\epsilon \sim \mathcal{N}(0, \sigma^2)\\
\end{equation}
According to Bayes theorem, if the prior distribution of $\tilde{\bm{w}}$ and $\sigma^2$ is conjugate to the data's likelihood function, the posterior probability distributions can be derived analytically. And then Thompson Sampling, as known as Posterior Sampling, is able to tackles the E\&E dilemma by maintaining the posterior over models and selecting creatives in proportion to the probability that they are optimal. We model the prior joint distribution of $\tilde{\bm{w}}$ and $\sigma^2$ as:
\begin{equation}\label{pi}
\begin{split}
\pi(\tilde{\bm{w}}, \sigma^{2}) &= \pi(\tilde{\bm{w}} | \sigma^{2}) \pi(\sigma^{2}),\\
\sigma^2 \sim  IG(a, b)  \ &and\   \tilde{\bm{w}} | \sigma^{2} \sim \mathcal{N} (\mu, \sigma^{2}\Sigma^{-1})\\
\end{split}
\end{equation}
where the $IG(\cdot)$ is an Inverse Gamma whose prior hyperparameters are set to $a_0 = b_0 = \eta>1$ and $\mathcal{N}(\cdot)$ is a Gaussian distribution with the 
initial parameters $\Sigma_0=\lambda Id$. Note that $\mu_0$ is initialized as the learned weights $\bm{w}$ of VAM in Equation \ref{equ5}. It can provide a better prior hyperparameters that further enhance the performance in the cold-starting phase. We call it VAM-Warmup and the results is shown in Figure \ref{overall_curve}(b).

Because we have chosen a conjugate prior, the posterior at time $t$ can be derived as
\begin{equation}\label{posterior}
\begin{split}
\Sigma(t) &= \bm{f}^{T}\bm{f} + \Sigma_0  \\
\mu(t) &= \Sigma(t)^{-1}(\Sigma_0\mu_0 + \bm{f}^{T}\bm{y})\\
a(t) &= a_0 + t/2 \\
b(t) &= b_0+\frac{1}{2}(\bm{y}^{T}\bm{y} + \mu_0^T\Sigma_0\mu_0 - \mu(t)^T\Sigma(t)\mu(t))
\end{split}
\end{equation}
where $\bm{f} \in \mathbb{R}^{t\times d}$  is a matrix that contain the content features for previous impressions and $\bm{y} \in \mathbb{R}^{t\times1}$ is the feedback rewards.
After updating the above parameters at $t$-th impression, we obtain the weight distributions with uncertainty estimation. We draw the weights $\bm{w}(t)$ from the learned distribution $\mathcal{N} (\mu(t), \sigma(t)^{2}\Sigma(t)^{-1})$ and select the best creative for product $I^n$ as 
\begin{equation}\label{decision}
C^n = \argmax_{ c\in \{C_1^n, C_2^n, \cdots, C_M^n\}} (\mathcal{N}_{feat}(c))^T\bm{w}(t) \\
\end{equation}

\textbf{The above model makes the weight distributions shared by all the products.} This simple linear assumption works well for small datasets, but becomes inferior when dealing with industrial data. For example, bright and vivid colors will be more attractive for women's top while concise colors are more proper for 3C digital accessories. In addition to this product-wise characteristic, a creative may contain a unique designed attribute that is not expressed by the shared weights. Hence, it is helpful to have weights that have both shared and non-shared components. 

We extend the Equation \ref{equ13} to the following hybrid model by combining product-wise and creative-specific linear terms. For creative $C_m^n$, it can be formulated as
\begin{equation}\label{hybrid-1}
y_m^n = \bm{f}_m^{nT}\bm{w}^n + \bm{f}_m^{nT}\bm{w}^n_m\\
\end{equation}
where $\bm{w}^n$ and $\bm{w}^n_m$ are product-wise and creative-specific parameters, and they are disjoinly optimized by Equation \ref{posterior}. Furthermore, we propose an fusion strategy to adaptively combine these two terms instead of the simple addition
\begin{equation}\label{hybrid-2}
y_m^n = (1-\lambda)\bm{f}_m^{nT}\bm{w}^n + \lambda\bm{f}_m^{nT}\bm{w}^n_m\\
\end{equation}
where $\lambda=(1+e^{\frac{-impression(I^n)+\theta_2}{\theta_1}})^{-1}$ is a sigmoid function with rescale $\theta_1$ and offset $\theta_2$. We find that if the impressions are inadequate, the product-wise parameters are learned better because it make use of the knowledge among all candidate creatives. Otherwise, the creative-specific term outperforms the shared one due to the sufficient feedback observations. The above procedure is shown in Algorithm ~\ref{algorithm1}. Because our hybrid model updates the parameters of each product independently, we take $I^n$ as example and adopt $(a^n(\cdot),\!b^n(\cdot),\!\mu^n(\cdot),\!\Sigma^n(\cdot))$ and $(a^n_m(\cdot),\!b^n_m(\cdot),\!\mu^n_m(\cdot),\!\Sigma^n_m(\cdot))$ to represent the shared and specific parameters.
\setlength{\intextsep}{0pt}
\begin{algorithm}[t]
  \caption{Hybrid Bandit Model}
  \label{algorithm1}
  \KwIn{$T>0$, product $I^n$, visual representations of candidate creatives $\bm{f}^n_0, \bm{f}^n_1, \cdots, \bm{f}^n_m, \cdots, \bm{f}^n_M$}
  Initialize the $a_0, b_0, \mu_0$ and $\Sigma_0$\;
  $a^n(0) \gets a_0, b^n(0) \gets b_0, \mu^n(0) \gets \mu_0, \Sigma^n(0)\gets\Sigma_0$\;
  $a^n_m(0) \gets a_0, b^n_m(0) \gets b_0, \mu^n_m(0) \gets \mu_0, \Sigma^n_m(0)\gets\Sigma_0$\;
  \For{$t=1, 2, 3, \dots, T$}
  {

    $\lambda = (1+e^{\frac{-impression(I^n)+\theta_2}{\theta_1}})^{-1}$\;
    \For{$m=1, 2, 3, \dots, M$}
    {
    	Sample ${\sigma^n}^2$ from $IG(a^n(t-1), b^n(t-1))$\;
	Sample $\bm{w}^n$ from $\mathcal{N} (\mu^n(t-1), {\sigma^n}^2\Sigma^n(t-1)^{-1})$\;
	Sample ${\sigma^n_m}^2$ from $IG(a^n_m(t-1), b^n_m(t-1))$\;
	Sample $\bm{w}^n_m$ from $\mathcal{N} (\mu^n_m(t-1), {\sigma^n_m}^2\Sigma^n_m(t-1)^{-1})$\;
    	$y^n_m =(1-\lambda)\bm{f}_m^{nT}\bm{w}^n + \lambda\bm{f}_m^{nT}\bm{w}^n_m$\;
    }
    
    $k = \argmax{(y^n_1, \dots, y^n_m, \dots, y^n_M)}$\;
    Display the creative $C^n_k$, and get the reward\;
    Update $a^n(t), b^n(t), \mu^n(t), \Sigma^n(t)$ by the historical data of product $I^n$ and Equation \ref{posterior}\;
    Update $a^n_k(t), b^n_k(t), \mu^n_k(t), \Sigma^n_k(t)$ by the historical data of creative $C^n_k$ and Equation \ref{posterior}\;
    Set the other parameters of time $t$ as the same as previous time $(t-1)$\;
    $impression(I^n) \leftarrow impression(I^n) + 1$\;
  }
\end{algorithm}
The distributions describe the uncertainty in weights which is related to impressed number: if there is less data, the model relies more on the visual evaluation results; Otherwise, the likelihood will reduce the priori effect so as to converge to the observation data. In order to fit the complex industrial data, we extend the shared linear model to the hybrid version, which consider both product-level knowledge and creative-specific information, and fused by empirical attention weights.



\section{Experiments}
\subsection{Dataset preparation and Evaluation Metrics}
\textbf{Dataset Preparation.} 
The description of CreativeRanking data is presented in Section\ref{data_collection}. 
The original images and rewards for each creative are provided in the order of displaying. For VAM, we aggregate the number of impressions and clicks to produce $\hat{CTR}$ by Equation \ref{equ12} on training set, and train the VAM using the loss function in Equation \ref{equ10}. For HBM, we update the policy by providing the visual representations extracted by VAM and the impression data like Equation \ref{data}. Note the interaction and policy updating procedure (see Algorithm ~\ref{algorithm1}) of HBM is conducted on the test set for simulating the online situations. We record the sequential interactions and rewards to measure the performance (see Algorithm ~\ref{algorithm2} and Equation \ref{hybrid-2}). Validation is used for hyperparameter tuning.

In addition to the CreativeRanking data, we also evaluate the methods on a public dataset, called Mushroom. Since there is no public dataset for creative ranking yet, we test the proposed hybrid bandit model on this standard dataset. The Mushroom Dataset \cite{schlimmer1981mushroom} contains 22 attributes for each mushroom, and two categories: poisonous and safe. Eating a safe mushroom will receive reward $+5$ while eating a poisonous one delivers reward $+5$ with probability $50\%$ and reward $-35$ otherwise. Not eating will provide no reward. We follow the protocols in \cite{riquelme2018deep}, and interact for 50000 rounds. 

\noindent{\textbf{Evaluation Metrics.}} For CreativeRanking data, we present two evaluation metrics to measure the performance, named simulated CTR ($sCTR$) and cumulative regret ($Regret$), respectively.

\noindent{\textbf{Simulated CTR ($sCTR$)}} is a practical metric which is quite close to the online performance. The details are shown in Algorithm ~\ref{algorithm2}. It replays the recorded impression data for all products. For each product, the policy will play $T^n$ rounds by receiving the recorded data $(C, y)$, and selects the best creative according to the predicted scores. If the selected one is the same as the $C$, the impression number, click number and policy itself will be updated (see line 3 to 14 in Algorithm ~\ref{algorithm2}).

\begin{algorithm}[b]
\small
  \caption{Evaluation Metrics - $sCTR$}
  \label{algorithm2}
  \KwIn{impression data, policy $\pi$}
  \KwOut{$sCTR$}
  $impressions \gets 0$\;
  $clicks \gets 0$\;
  \For{$n=1, 2, 3, \dots, N$}
  {
    \For{$t=1, 2, 3, \dots, T^n$}
     {
  	Get next impression (C, y)\;
	Get predicated scores $(y^n_1, \dots, y^n_M)$ by policy $\pi$\;
	$k = \argmax{(y^n_1, \dots, y^n_M)}$\;
	\If{$C^n_k = C$}
	  {
	    $impressions\gets impressions+1$\;
	    $clicks \gets clicks+y$\;
	    update policy $\pi$ by data (C, y)\;
	  }
    }
  }
  $sCTR = \frac{clicks}{impressions}$\;
  return $sCTR$
\end{algorithm}

Take HBM as an example, algorithm ~\ref{algorithm1} shows the online update process. To test the HBM by using offline data, we can change the action ``display and update'' (line 14 to 18 in Algorithm ~\ref{algorithm1}) to the conditioned version in the line 8 to 12 in Algorithm~\ref{algorithm2}.

\textbf{Cumulative regret ($Regret$)} is commonly used for evaluating bandit models. It is defined as 
\begin{equation}\label{regret-1}
Regret = E[r^* - r]\\
\end{equation}
where $r^*$ is the cumulative reward of the optimal policy, i.e., the policy that always selects the action with highest expected reward given the context \cite{riquelme2018deep}. Specifically, we select the optimal creative for our dataset, and calculate the $Regret$ as 
\begin{equation}\label{equ_regret}
Regret = \frac{\sum_{n=1}^{N}{click(C^n)}}{\sum_{n=1}^{N}{impression(C^n)}} - sCTR\\
\end{equation}
where $sCTR$ should be produced by Algorithm ~\ref{algorithm2} first. And the $C^n$ is selected by calculating $\hat{CTR}$ in Equation \ref{equ2} on the test set. 

For Mushroom, we follow the definition of cumulative regret in \cite{riquelme2018deep} to evaluate the models. 

\subsection{Implementation details}
The model was implemented with Pytorch \cite{paszke2017automatic}. We adopt deep residual network (ResNet-18)\cite{he2016deep} pretrained on ImageNet classification \cite{deng2009imagenet} as backbone, and the model is finetuned with Creative Ranking task. For VAM, we use stochastic gradient descent (SGD) with a mini-batch of 64 per GPU. The learning rate is initially set to 0.01 and then gradually decreased to $e-4$. The training process lasts 30 epochs on the datasets. For HBM, we extract the feature representations $\bm{f}_m^n$ from VAM, and update the weights distribution $\bm{w}_m^n$ and  $\bm{w}^n$ by using bayesian regression.

\subsection{Comparison with State-of-the-art Systems}
In this subsection, we show the performance of the related methods in Table \ref{overall_performance} and Figure \ref{overall_curve}. The methods are divided into some groups: a uniform strategy, context-free bandit models,  linear bandit models, neural bandit models and our proposed methods. Table \ref{overall_performance} presents the   $Regret$ and $sCTR$ of all above models on both Mushroom and CreativeRanking datasets, and our methods - (NN/VAM-HBM) exhibits state-of-the-art results compared to the related models. We also conduct further analysis by showing the reward tendency of consecutive 15 days in Figure \ref{overall_curve}. Daily $sCTR$ evaluates the model for each day independently, showing the flexibility of the policy when interacting with the feedback. And cumulative $sCTR$ presents the cumulative rewards up to the specific day which is used to measure the overall performance.

\textbf{\emph{Uniform}}: The baseline strategy that randomly selects an action (eat/not eat for Mushroom and one creative for CreativeRanking). Because this strategy has neither prior knowledge nor abilities of learning from the data, it gets poor performance on the test sets. 

\textbf{Context-free Bandit Models}: Epsilon-greedy \cite{sutton2018reinforcement}, Thompson sampling \cite{russo2014learning} and Upper Confidence Bounds (UCB) approaches \cite{auer2002finite} are simple yet effective strategies to deal with the bandit problem. They rely on history impression data (click/non-click) and keep updating their strategies. However, for the cold-start stage, they might randomly choose a creative like ``Uniform'' strategy (orange lines in Figure \ref{overall_curve}(c) in the first few days). We find that their curves are gradually rising, but without prior information, the overall performance is inferior to the other models.

\textbf{Linear Bandit Models}:  The linear bandit model is an extension to the context-free method by incorporating contextual information. For Mushroom, we adopt the 22 attributes to describe a mushroom, such as shape, color and so on. The $Regret$ is reduced when combining the side information. For CreativeRanking, we use color distribution \cite{azimi2012impact} to represent a creative, and update the linear payoff functions. From the results in Table \ref{overall_performance}, the linear models achieve better results than the context-free methods, but they still face the problem of lacking representational power. 

\textbf{Neural Bandit Models}: The neural bandit models add a linear regression on top of the neural network. In Table \ref{overall_performance}, ``NN'' denotes fully connected layers that used for extracting mushroom representations. For CreativeRanking, all these neural models use our VAM as feature extractor, and adopt different E\&E policies. Figure \ref{overall_curve}(a) reveals some interesting observations: (1) The orange and blue lines represent the  E-greedy and VAM-Greedy, respectively. With the visual priors, VAM-Greedy achieves much better performance at the beginning (about 5\% CTR lift), which demonstrates the effectiveness of the visual evaluation. (2) Because VAM-Greedy is a greedy strategy that lack of exploration, it becomes mediocre in the long run. When incorporating E\&E model - HBM, our VAM-HBM outperforms the other baselines by a significant margin. Besides, we also use Dropout as a Bayesian approximation\cite{gal2016dropout}, but it is not able estimate the uncertainty as accurate as the other policies.

\textbf{Our Methods}: We propose VAM-Warmup that initialize the $\mu_0$ in bandit model by learned weights in VAM. By comparing red and blue dashed lines in Figure \ref{overall_curve}(b), we find the parameters with prior distributions improves 1.7\% CTR for overall performance. In addition, we extend the model by adding creative-specific parameters, named VAM-HBM, and it further enhances the model capacity and achieves the state-of-the-art result, especially the impressions for creatives become adequate (see solid red line in Figure \ref{overall_curve}(b)(c)(d)). For Mushroom dataset, in order to demonstrate the idea, we cluster the data into 2 groups by attribute ``bruises'', each maintaining the individual parameters. When combining the individual and shared parameters by fusion weights in Equation \ref{hybrid-2}, the model reduces the $Regret$ to 1.93. Note that we use the default hyperparameters provided by NeuralLinear without carefully tuning.

\begin{table*}[t]
\vspace{0.2cm}
\setlength{\abovecaptionskip}{0cm} 
\begin{minipage}{0.48\textwidth} 
\centering 
\arrayrulecolor{black} 
\renewcommand\arraystretch{1.4}
\begin{tabular}{p{3.3cm}p{1.3cm}p{1.3cm}p{1.2cm}}
\hline
 & Mushroom & \multicolumn{2}{c}{CreativeRanking} \\
\hline
 Evaluation Metrics& {Regret (\%)} & Regret (\%) & sCTR (\%)\\
\hline
Uniform&100&100&2.950\\
\hline
\multicolumn{4}{l}{Context-free Bandit Models (Orange lines)}\\
\arrayrulecolor{gray} \cdashline{1-4}[0.8pt/2pt]
E-Greedy\cite{sutton2018reinforcement}&52.99&87.22&3.166\\
Thompson Sampling\cite{russo2014learning}&52.49&87.69&3.158\\
UCB\cite{auer2002finite}&52.42&87.04&3.169\\
\arrayrulecolor{black} 
\hline
\multicolumn{4}{l}{Linear Bandit Models (Green lines)}\\
\arrayrulecolor{gray} \cdashline{1-4}[0.8pt/2pt]
LinGreedy\cite{riquelme2018deep}&14.28&91.72&3.090\\
LinThompson\cite{riquelme2018deep}&2.37&85.68&3.192\\
LinUCB\cite{li2010contextual}&10.27&85.50&3.195\\
\arrayrulecolor{black} 
\hline
\multicolumn{4}{l}{Neural Bandit Models (Blue lines)}\\
\arrayrulecolor{gray} \cdashline{1-4}[0.8pt/2pt]
NN/VAM-Greedy&6.68&84.11&3.219\\
NN/VAM-Thompson\cite{riquelme2018deep}&2.22&83.02&3.237\\
NN/VAM-UCB&7.51&83.91&3.222\\
NN/VAM-Dropout\cite{gal2016dropout}&5.57&84.32&3.215\\
\arrayrulecolor{black} 
\hline
\multicolumn{4}{l}{Our Methods (red lines)}\\
\arrayrulecolor{gray} \cdashline{1-4}[0.8pt/2pt]
VAM-Warmup &-&79.70&3.293\\
\textbf{NN/VAM-HBM}&\textbf{1.93}&\textbf{78.11}&\textbf{3.320}\\
\arrayrulecolor{black} 
\hline
\end{tabular}  
\caption{Performance comparison with state-of-the-art systems on both \textit{Mushroom} and \textit{CreativeRanking} test set. $Regret$ is Normalized with respect to the performance of Uniform.} 
\label{overall_performance}
\end{minipage}%
\hspace{0.1in}
\begin{minipage}{0.48\textwidth} 
\setlength{\abovecaptionskip}{0.5cm}
\centering 
\includegraphics[width=1\textwidth]{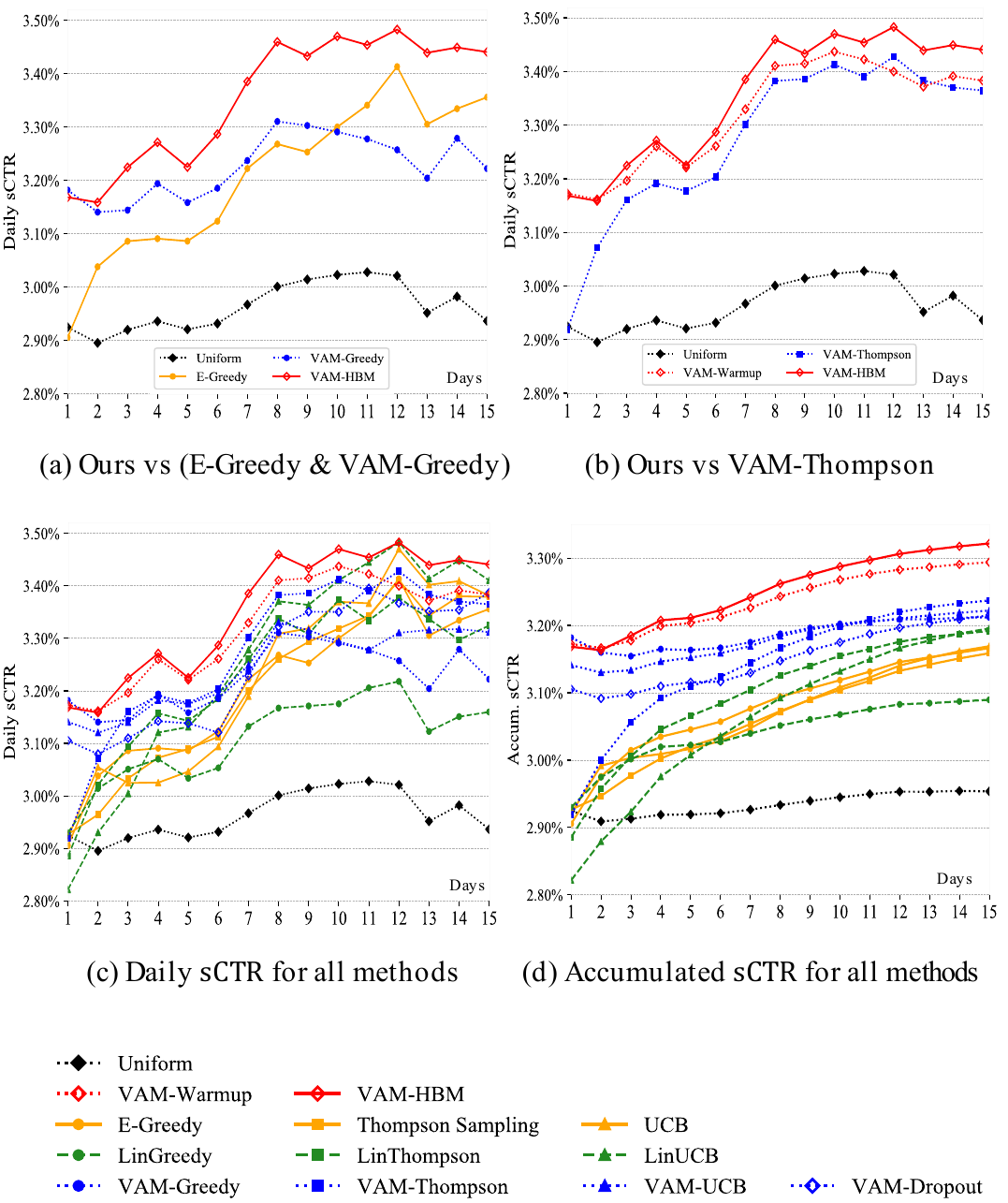} 
\captionof{figure}{Reward tendency of consecutive 15 days on \textit{CreativeRanking}.}
\label{overall_curve} 
\end{minipage}%
\end{table*}


\begin{table}
\setlength{\abovecaptionskip}{-0.3cm}
\centering 
\small
\renewcommand\arraystretch{1.4}
\begin{tabular}{p{2.2cm}|p{0.6cm}|p{0.9cm}|p{0.9cm}|p{0.9cm}|p{0.9cm}} 
\hline
Methods &Base & (a)&(b)&(c)& (d)\\
\hline
Point-wise Loss?   &&{${\surd}$} &&{${\surd}$}&{${\surd}$}  \\
List-wise Loss?   &&&{${\surd}$}&{${\surd}$}&{${\surd}$}\\
Noise Mitigation? &&&&&{${\surd}$}\\
\hline
$sCTR(\%)$  &2.950&3.140\blank{0.3cm}$\uparrow6.4\%$&3.167\blank{0.2cm}$\uparrow 7.4\%$&3.194\blank{0.2cm}$\uparrow 8.3\%$&3.219\blank{0.2cm}$\uparrow 9.1\%$\\
\hline
\end{tabular}
\vspace{0.5em}
\caption{Ablation study for each component in the VAM. $sCTR$ are performed on the \textit{CreativeRanking} test set and $\uparrow sCTR$ lift is calculated by $\frac{(sCTR(*)-sCTR(base))}{sCTR(base)}*100\%$.}
\label{ablation}    
\end{table}
\subsection{Ablation Study}

In this subsection, we conduct an ablation study on CreativeRanking dataset so as to validate the effectiveness of each component in the VAM, including list-wise ranking loss, point-wise auxiliary regression loss and noise mitigation. Besides,  we also compare our VAM with ``learning-to-rank'' visual models (including aesthetic models).  We show the results  in Table \ref{ablation} and Table \ref{loss} to demonstrate the consistent improvements.

\textbf{\emph{Base}} in Table \ref{ablation} stands for the baseline result. We adopt ``uniform'' strategy that randomly choose a creative among the candidates. The baseline is 2.950\% for $sCTR$.

\textbf{\emph{Method (a) and (b)}}: Method (a) and (b) utilize point-wise (Equation \ref{equ9}) and list-wise loss (Equation \ref{equ8}) as the objective function, respectively. Although the model has never seen the products/creatives on the test set before, it has learned general patterns to identify more attractive creatives. Moreover, the ranking loss concentrates on the top-1 probability learning which is more suitable than the point-wise objective for our scenarios. The simple version (b) can improve the $sCTR$ by $7.4\%$.

\textbf{\emph{Method (c)}}: Method (c) combines the point-wise auxiliary regression loss with the ranking objective. It not only learns the relative order of creative quality, but also the absolute CTRs. We find it is good at fitting the real CTR distributions and achieve the better performance 3.194\% (8.3\% lift) for $sCTR$. 

\textbf{\emph{Method (d)}}: Method (d) contains \emph{label smooth} and \emph{weighted sampler}, both of which are designed for mitigating the label noise. Weighted sampler makes the model pay more attention to the samples whose impression numbers are sufficient while label smooth aims to improve the label reliability. These two practical methods further improve the $sCTR$ to 3.216\%, lifting 9.1\% in total.

\begin{table}
\small
\setlength{\abovecaptionskip}{-0.3cm}
\centering 
\renewcommand\arraystretch{1.4}
\begin{tabular}{p{4cm}|p{2cm}} 
\hline
Ranking Loss & sCTR (\%)\\
\hline
Pairwise Hinge Loss \cite{gattupalli2016computational} & 3.170 \\
Aesthetics Ranking Loss \cite{kong2016photo}& 3.167 \\
Triplet Loss \cite{schwarz2018will} & 3.115\\
Pairwise \cite{zhao2019you} & 3.188\\
\hline
VAM (Ours) & \textbf{3.219}\\
\hline
\end{tabular}
\vspace{0.5em}
\caption{Comparison with other ``learn-to-rank'' visual models. All above models adopt ResNet-18 as backbone.}
\label{loss}   
\end{table}

\textbf{\emph{Related Loss functions}}: 
Pair-wise and triplet loss are typical loss functions for learning to rank problems. \cite{gattupalli2016computational,kong2016photo,schwarz2018will} adopt hinge loss that is used for "maximum-margin" classification between the better candidate and the other one. It only requires the better creative to get higher score than the other one by a pre-defined margin, without consideration of the exact difference. Our loss function in Equation \ref{equ8} and \ref{equ9} estimate their CTR gaps and produce more accurate differences. \cite{zhao2019you} employ \cite{BurgesSRLDHH05} as their pair-wise framework.  Compared to the pair-wise learning, we treat one product as a training sample and use list of creatives as instances. It is more efficient and suitable with real scenarios which will display the best creative for one impression. Thus, our method obtains the leading performance on $sCTR$.

In summary, the proposed list-wise method enables the model focus on learning creative qualities and obtains better generalizability. Incorporating point-wise regression and noise mitigation techniques is able to enhance the model capacity of fitting the real-world data.

\begin{figure*}[t]
\setlength{\belowcaptionskip}{-0.1cm}
\setlength{\abovecaptionskip}{0.1cm}
\begin{center}
\includegraphics[width=1\linewidth]{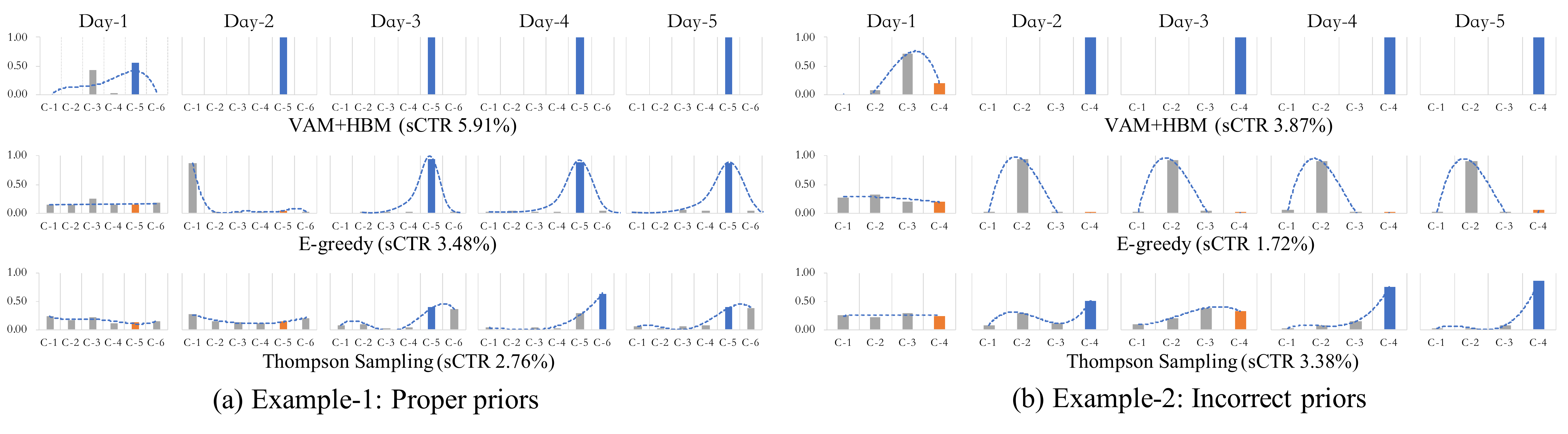}
\end{center}
\caption{Two typical cases that present the changing of strategies. The horizontal axis shows different creatives while the vertical axis is the probability of being displayed for creatives. ``Proper priors'' indicates that VAM provides right predictions and ``Incorrect priors'' otherwise.}
\label{case}
\end{figure*}

\subsection{Hyperparameter Settings}

\noindent{\textbf{$\gamma$ in Equation \ref{equ10}.}} We tune hyperparameters in the validation set. $\gamma$ in Equation \ref{equ10} is adopted to control the weight of point-wise auxiliary loss. According to the validation results (see Table \ref{gamma}), we take $\gamma=0.5$. It is consistent with our hypothesis that ranking loss should play a more important role in the creative ranking task.

\begin{table}
\setlength{\abovecaptionskip}{-0.3cm}
\centering 
\small
\renewcommand\arraystretch{1.4}
\begin{tabular}{p{2.2cm}|p{0.6cm}p{0.6cm}p{0.6cm}p{0.6cm}p{0.6cm}} 
\hline
$\gamma$ in Equation \ref{equ10} &0.0 & 0.1 & 0.5 & 1.0 & 2.0\\
\hline
Validation sCTR(\%) &3.15& 3.15 &3.17& 3.16& 3.13\\
Test sCTR(\%) &3.17&3.19& 3.22& 3.18 &3.15\\
\hline
\end{tabular}
\vspace{0.5em}
\caption{Val/Test $sCTR$ with different $\gamma$ in Equation \ref{equ10}. }
\label{gamma}    
\end{table}
\begin{table}
\setlength{\abovecaptionskip}{-0.3cm}
\centering 
\small
\renewcommand\arraystretch{1.4}
\begin{tabular}{p{2.0cm}|p{1.4cm}p{1.4cm}p{1.4cm}} 
\hline
$\theta_1 / \theta_2$ in $\lambda$ &125 & 150 & 175 \\
\hline
30 &3.27\%(3.32\%)& 3.28\%(3.33\%) &3.28\%(3.31\%)\\
50 &3.27\%(3.31\%) &3.28\%(3.32\%) &3.27\%(3.32\%)\\
100& 3.27\%(3.31\%) &3.27\%(3.31\%) &3.27\%(3.31\%)\\
\hline
\end{tabular}
\vspace{0.5em}
\caption{x\%(x\%) denotes val(test) $sCTR$ of different $\theta_1 / \theta_2$ in $\lambda$.}
\label{theta}    
\end{table}

\noindent{\textbf{$\theta_1 / \theta_2$ of $\lambda$ in Equation \ref{hybrid-2}.}} $\theta_1/ \theta_2$ control the rescale and offset of $\lambda$ in Equation \ref{hybrid-2}. Optimal hyperparameters vary in different real-world platforms(e.g., offset is set to 150, around the mean impression number of each creative). We find the final performance is not sensitive to these hyperparameters (see Table \ref{theta}). We choose $\theta_1=50$ and $\theta_2=150$ in our experiments.

\subsection{Case Study}

\begin{figure}[t]
\setlength{\abovecaptionskip}{0.3cm}
\begin{center}
\includegraphics[width=1\linewidth]{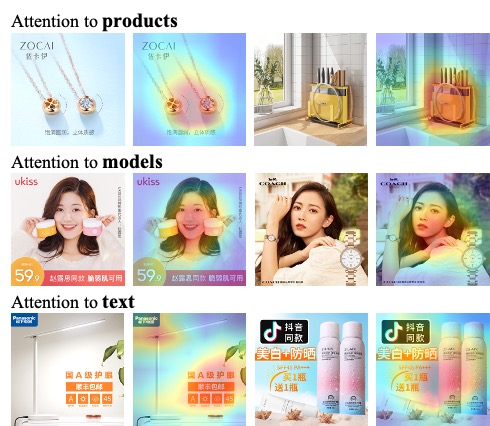}
\end{center}
\caption{Visualization of the learned VAM. The model pays attention to different regions adaptively, including products, models and the text on the creative.}
\label{cnn_vis}
\end{figure}
\noindent{\textbf{Strategy Visualization.}} 
We show two typical cases that exhibit the changing of strategies. Figure \ref{case} (a) shows the proper prior of HBM. We believe that the best creative should have the largest displaying probability among candidates. If this expectation is satisfied, a blue bar is shown; otherwise, orange bars are shown. It grants most impression opportunities to creative C-5 from the first day, while the other two methods spend 2 days to find the best creative.  For another case that receives incorrect prior in Figure \ref{case}(b), the HBM adjusts the decision by considering the online feedback. The interactions help to revise the prior knowledge and fit to the real-world feedback. Form this comparison, we find the HBM makes good use of visual priors, and adjusts flexibly according to the feedback signals.

\noindent{\textbf{CNN Visualization.}}
Besides ranking performance, we would like to attain further insight into the learned VAM. To this end, we visualize the response of our VAM according to the activations on the high-level feature maps, and the resulting visualization is shown in Figure \ref{cnn_vis}. By learning from the creative ranking, we find that the CNN pays attention to different regions adaptively, including products, models and the text on the creative. As shown in the second row Figure \ref{cnn_vis}, the VAM draw higher attention to the models rather than the products. It may caused by the reason that products endorsed by celebrities are more attractive than simply displaying the products. Besides, some textual information, such as description and discount information, can also attract customers.

\section{Conclusions}
In this paper, we propose a hybrid bandit model with visual priors. To the best of our knowledge, this is the first time that formulates the creative ranking as a E\&E problem with visual priors.
The VAM adopts a list-wise ranking loss function for ordering the creative quality only by their contents.  In addition to the ability of visual evaluation, we extend the model to be updated when receiving feedback from online scenarios called HBM. Last but not the least, we construct and release a novel large-scale creative dataset named \textit{CreativeRanking}. We would like to  draw more attention to this topic which benefits both the research community and website’s user experience. We carried out extensive experiments, including performance comparison, ablation study and case study, demonstrating the solid improvements of the proposed model.


\bibliographystyle{ACM-Reference-Format}
\bibliography{sample-base}


\begin{thebibliography}{34}


\ifx \showCODEN    \undefined \def \showCODEN     #1{\unskip}     \fi
\ifx \showDOI      \undefined \def \showDOI       #1{#1}\fi
\ifx \showISBNx    \undefined \def \showISBNx     #1{\unskip}     \fi
\ifx \showISBNxiii \undefined \def \showISBNxiii  #1{\unskip}     \fi
\ifx \showISSN     \undefined \def \showISSN      #1{\unskip}     \fi
\ifx \showLCCN     \undefined \def \showLCCN      #1{\unskip}     \fi
\ifx \shownote     \undefined \def \shownote      #1{#1}          \fi
\ifx \showarticletitle \undefined \def \showarticletitle #1{#1}   \fi
\ifx \showURL      \undefined \def \showURL       {\relax}        \fi
\providecommand\bibfield[2]{#2}
\providecommand\bibinfo[2]{#2}
\providecommand\natexlab[1]{#1}
\providecommand\showeprint[2][]{arXiv:#2}

\bibitem[\protect\citeauthoryear{Agrawal and Goyal}{Agrawal and Goyal}{2013}]%
        {agrawal2013thompson}
\bibfield{author}{\bibinfo{person}{Shipra Agrawal} {and} \bibinfo{person}{Navin
  Goyal}.} \bibinfo{year}{2013}\natexlab{}.
\newblock \showarticletitle{Thompson Sampling for Contextual Bandits with
  Linear Payoffs}. In \bibinfo{booktitle}{\emph{Proceedings of the 30th
  International Conference on Machine Learning, {ICML} 2013, Atlanta, GA, USA,
  16-21 June 2013}} \emph{(\bibinfo{series}{{JMLR} Workshop and Conference
  Proceedings}, Vol.~\bibinfo{volume}{28})}. \bibinfo{publisher}{JMLR.org},
  \bibinfo{pages}{127--135}.
\newblock
\urldef\tempurl%
\url{http://proceedings.mlr.press/v28/agrawal13.html}
\showURL{%
\tempurl}


\bibitem[\protect\citeauthoryear{Auer, Cesa{-}Bianchi, and Fischer}{Auer
  et~al\mbox{.}}{2002}]%
        {auer2002finite}
\bibfield{author}{\bibinfo{person}{Peter Auer}, \bibinfo{person}{Nicol{\`{o}}
  Cesa{-}Bianchi}, {and} \bibinfo{person}{Paul Fischer}.}
  \bibinfo{year}{2002}\natexlab{}.
\newblock \showarticletitle{Finite-time Analysis of the Multiarmed Bandit
  Problem}.
\newblock \bibinfo{journal}{\emph{Mach. Learn.}} \bibinfo{volume}{47},
  \bibinfo{number}{2-3} (\bibinfo{year}{2002}), \bibinfo{pages}{235--256}.
\newblock
\urldef\tempurl%
\url{https://doi.org/10.1023/A:1013689704352}
\showDOI{\tempurl}


\bibitem[\protect\citeauthoryear{Azimi, Zhang, Zhou, Navalpakkam, Mao, and
  Fern}{Azimi et~al\mbox{.}}{2012}]%
        {azimi2012impact}
\bibfield{author}{\bibinfo{person}{Javad Azimi}, \bibinfo{person}{Ruofei
  Zhang}, \bibinfo{person}{Yang Zhou}, \bibinfo{person}{Vidhya Navalpakkam},
  \bibinfo{person}{Jianchang Mao}, {and} \bibinfo{person}{Xiaoli~Z. Fern}.}
  \bibinfo{year}{2012}\natexlab{}.
\newblock \showarticletitle{The impact of visual appearance on user response in
  online display advertising}. In \bibinfo{booktitle}{\emph{Proceedings of the
  21st World Wide Web Conference, {WWW} 2012, Lyon, France, April 16-20, 2012
  (Companion Volume)}}, \bibfield{editor}{\bibinfo{person}{Alain Mille},
  \bibinfo{person}{Fabien~L. Gandon}, \bibinfo{person}{Jacques Misselis},
  \bibinfo{person}{Michael Rabinovich}, {and} \bibinfo{person}{Steffen Staab}}
  (Eds.). \bibinfo{publisher}{{ACM}}, \bibinfo{pages}{457--458}.
\newblock
\urldef\tempurl%
\url{https://doi.org/10.1145/2187980.2188075}
\showDOI{\tempurl}


\bibitem[\protect\citeauthoryear{Burges, Shaked, Renshaw, Lazier, Deeds,
  Hamilton, and Hullender}{Burges et~al\mbox{.}}{2005}]%
        {BurgesSRLDHH05}
\bibfield{author}{\bibinfo{person}{Christopher J.~C. Burges},
  \bibinfo{person}{Tal Shaked}, \bibinfo{person}{Erin Renshaw},
  \bibinfo{person}{Ari Lazier}, \bibinfo{person}{Matt Deeds},
  \bibinfo{person}{Nicole Hamilton}, {and} \bibinfo{person}{Gregory~N.
  Hullender}.} \bibinfo{year}{2005}\natexlab{}.
\newblock \showarticletitle{Learning to rank using gradient descent}. In
  \bibinfo{booktitle}{\emph{Machine Learning, Proceedings of the Twenty-Second
  International Conference {(ICML} 2005), Bonn, Germany, August 7-11, 2005}}
  \emph{(\bibinfo{series}{{ACM} International Conference Proceeding Series},
  Vol.~\bibinfo{volume}{119})}, \bibfield{editor}{\bibinfo{person}{Luc~De
  Raedt} {and} \bibinfo{person}{Stefan Wrobel}} (Eds.).
  \bibinfo{publisher}{{ACM}}, \bibinfo{pages}{89--96}.
\newblock


\bibitem[\protect\citeauthoryear{Cao, Qin, Liu, Tsai, and Li}{Cao
  et~al\mbox{.}}{2007}]%
        {cao2007learning}
\bibfield{author}{\bibinfo{person}{Zhe Cao}, \bibinfo{person}{Tao Qin},
  \bibinfo{person}{Tie{-}Yan Liu}, \bibinfo{person}{Ming{-}Feng Tsai}, {and}
  \bibinfo{person}{Hang Li}.} \bibinfo{year}{2007}\natexlab{}.
\newblock \showarticletitle{Learning to rank: from pairwise approach to
  listwise approach}. In \bibinfo{booktitle}{\emph{Machine Learning,
  Proceedings of the Twenty-Fourth International Conference {(ICML} 2007),
  Corvallis, Oregon, USA, June 20-24, 2007}} \emph{(\bibinfo{series}{{ACM}
  International Conference Proceeding Series}, Vol.~\bibinfo{volume}{227})},
  \bibfield{editor}{\bibinfo{person}{Zoubin Ghahramani}} (Ed.).
  \bibinfo{publisher}{{ACM}}, \bibinfo{pages}{129--136}.
\newblock
\urldef\tempurl%
\url{https://doi.org/10.1145/1273496.1273513}
\showDOI{\tempurl}


\bibitem[\protect\citeauthoryear{Capelo, Aggarwal, and Yadav}{Capelo
  et~al\mbox{.}}{2019}]%
        {capelo2019combining}
\bibfield{author}{\bibinfo{person}{Mark Capelo}, \bibinfo{person}{Karan
  Aggarwal}, {and} \bibinfo{person}{Pranjul Yadav}.}
  \bibinfo{year}{2019}\natexlab{}.
\newblock \showarticletitle{Combining Text and Image data for Product
  Recommendability Modeling}. In \bibinfo{booktitle}{\emph{2019 {IEEE}
  International Conference on Big Data (Big Data), Los Angeles, CA, USA,
  December 9-12, 2019}}. \bibinfo{publisher}{{IEEE}},
  \bibinfo{pages}{5992--5994}.
\newblock
\urldef\tempurl%
\url{https://doi.org/10.1109/BigData47090.2019.9006197}
\showDOI{\tempurl}


\bibitem[\protect\citeauthoryear{Chandakkar, Gattupalli, and Li}{Chandakkar
  et~al\mbox{.}}{2017}]%
        {gattupalli2016computational}
\bibfield{author}{\bibinfo{person}{Parag~S. Chandakkar},
  \bibinfo{person}{Vijetha Gattupalli}, {and} \bibinfo{person}{Baoxin Li}.}
  \bibinfo{year}{2017}\natexlab{}.
\newblock \showarticletitle{A Computational Approach to Relative Aesthetics}.
\newblock \bibinfo{journal}{\emph{CoRR}}  \bibinfo{volume}{abs/1704.01248}
  (\bibinfo{year}{2017}).
\newblock
\showeprint[arxiv]{1704.01248}
\urldef\tempurl%
\url{http://arxiv.org/abs/1704.01248}
\showURL{%
\tempurl}


\bibitem[\protect\citeauthoryear{Chen, Sun, Li, Lu, and Hua}{Chen
  et~al\mbox{.}}{2016}]%
        {chen2016deep}
\bibfield{author}{\bibinfo{person}{Junxuan Chen}, \bibinfo{person}{Baigui Sun},
  \bibinfo{person}{Hao Li}, \bibinfo{person}{Hongtao Lu}, {and}
  \bibinfo{person}{Xian{-}Sheng Hua}.} \bibinfo{year}{2016}\natexlab{}.
\newblock \showarticletitle{Deep {CTR} Prediction in Display Advertising}. In
  \bibinfo{booktitle}{\emph{Proceedings of the 2016 {ACM} Conference on
  Multimedia Conference, {MM} 2016, Amsterdam, The Netherlands, October 15-19,
  2016}}, \bibfield{editor}{\bibinfo{person}{Alan Hanjalic},
  \bibinfo{person}{Cees Snoek}, \bibinfo{person}{Marcel Worring},
  \bibinfo{person}{Dick C.~A. Bulterman}, \bibinfo{person}{Benoit Huet},
  \bibinfo{person}{Aisling Kelliher}, \bibinfo{person}{Yiannis Kompatsiaris},
  {and} \bibinfo{person}{Jin Li}} (Eds.). \bibinfo{publisher}{{ACM}},
  \bibinfo{pages}{811--820}.
\newblock
\urldef\tempurl%
\url{https://doi.org/10.1145/2964284.2964325}
\showDOI{\tempurl}


\bibitem[\protect\citeauthoryear{Cheng, van Zwol, Azimi, Manavoglu, Zhang,
  Zhou, and Navalpakkam}{Cheng et~al\mbox{.}}{2012}]%
        {ChengZAMZZN12}
\bibfield{author}{\bibinfo{person}{Haibin Cheng}, \bibinfo{person}{Roelof van
  Zwol}, \bibinfo{person}{Javad Azimi}, \bibinfo{person}{Eren Manavoglu},
  \bibinfo{person}{Ruofei Zhang}, \bibinfo{person}{Yang Zhou}, {and}
  \bibinfo{person}{Vidhya Navalpakkam}.} \bibinfo{year}{2012}\natexlab{}.
\newblock \showarticletitle{Multimedia features for click prediction of new ads
  in display advertising}. In \bibinfo{booktitle}{\emph{The 18th {ACM} {SIGKDD}
  International Conference on Knowledge Discovery and Data Mining, {KDD}}},
  \bibfield{editor}{\bibinfo{person}{Qiang Yang}, \bibinfo{person}{Deepak
  Agarwal}, {and} \bibinfo{person}{Jian Pei}} (Eds.).
  \bibinfo{publisher}{{ACM}}, \bibinfo{pages}{777--785}.
\newblock


\bibitem[\protect\citeauthoryear{Deng, Dong, Socher, Li, Li, and Li}{Deng
  et~al\mbox{.}}{2009}]%
        {deng2009imagenet}
\bibfield{author}{\bibinfo{person}{Jia Deng}, \bibinfo{person}{Wei Dong},
  \bibinfo{person}{Richard Socher}, \bibinfo{person}{Li{-}Jia Li},
  \bibinfo{person}{Kai Li}, {and} \bibinfo{person}{Fei{-}Fei Li}.}
  \bibinfo{year}{2009}\natexlab{}.
\newblock \showarticletitle{ImageNet: {A} large-scale hierarchical image
  database}. In \bibinfo{booktitle}{\emph{2009 {IEEE} Computer Society
  Conference on Computer Vision and Pattern Recognition {(CVPR} 2009), 20-25
  June 2009, Miami, Florida, {USA}}}. \bibinfo{publisher}{{IEEE} Computer
  Society}, \bibinfo{pages}{248--255}.
\newblock
\urldef\tempurl%
\url{https://doi.org/10.1109/CVPR.2009.5206848}
\showDOI{\tempurl}


\bibitem[\protect\citeauthoryear{Esfandarani and Milanfar}{Esfandarani and
  Milanfar}{2018}]%
        {talebi2018nima}
\bibfield{author}{\bibinfo{person}{Hossein~Talebi Esfandarani} {and}
  \bibinfo{person}{Peyman Milanfar}.} \bibinfo{year}{2018}\natexlab{}.
\newblock \showarticletitle{{NIMA:} Neural Image Assessment}.
\newblock \bibinfo{journal}{\emph{{IEEE} Trans. Image Process.}}
  \bibinfo{volume}{27}, \bibinfo{number}{8} (\bibinfo{year}{2018}),
  \bibinfo{pages}{3998--4011}.
\newblock
\urldef\tempurl%
\url{https://doi.org/10.1109/TIP.2018.2831899}
\showDOI{\tempurl}


\bibitem[\protect\citeauthoryear{Fran{\c{c}}ois{-}Lavet, Henderson, Islam,
  Bellemare, and Pineau}{Fran{\c{c}}ois{-}Lavet et~al\mbox{.}}{2018}]%
        {sutton2018reinforcement}
\bibfield{author}{\bibinfo{person}{Vincent Fran{\c{c}}ois{-}Lavet},
  \bibinfo{person}{Peter Henderson}, \bibinfo{person}{Riashat Islam},
  \bibinfo{person}{Marc~G. Bellemare}, {and} \bibinfo{person}{Joelle Pineau}.}
  \bibinfo{year}{2018}\natexlab{}.
\newblock \showarticletitle{An Introduction to Deep Reinforcement Learning}.
\newblock \bibinfo{journal}{\emph{Found. Trends Mach. Learn.}}
  \bibinfo{volume}{11}, \bibinfo{number}{3-4} (\bibinfo{year}{2018}),
  \bibinfo{pages}{219--354}.
\newblock
\urldef\tempurl%
\url{https://doi.org/10.1561/2200000071}
\showDOI{\tempurl}


\bibitem[\protect\citeauthoryear{Gal and Ghahramani}{Gal and
  Ghahramani}{2016}]%
        {gal2016dropout}
\bibfield{author}{\bibinfo{person}{Yarin Gal} {and} \bibinfo{person}{Zoubin
  Ghahramani}.} \bibinfo{year}{2016}\natexlab{}.
\newblock \showarticletitle{Dropout as a Bayesian Approximation: Representing
  Model Uncertainty in Deep Learning}. In \bibinfo{booktitle}{\emph{Proceedings
  of the 33nd International Conference on Machine Learning, {ICML} 2016, New
  York City, NY, USA, June 19-24, 2016}} \emph{(\bibinfo{series}{{JMLR}
  Workshop and Conference Proceedings}, Vol.~\bibinfo{volume}{48})},
  \bibfield{editor}{\bibinfo{person}{Maria{-}Florina Balcan} {and}
  \bibinfo{person}{Kilian~Q. Weinberger}} (Eds.).
  \bibinfo{publisher}{JMLR.org}, \bibinfo{pages}{1050--1059}.
\newblock
\urldef\tempurl%
\url{http://proceedings.mlr.press/v48/gal16.html}
\showURL{%
\tempurl}


\bibitem[\protect\citeauthoryear{Ge, Zhao, Zhou, Chen, Liu, Yi, Hu, Liu, Sun,
  Liu, et~al\mbox{.}}{Ge et~al\mbox{.}}{2018}]%
        {ge2018image}
\bibfield{author}{\bibinfo{person}{Tiezheng Ge}, \bibinfo{person}{Liqin Zhao},
  \bibinfo{person}{Guorui Zhou}, \bibinfo{person}{Keyu Chen},
  \bibinfo{person}{Shuying Liu}, \bibinfo{person}{Huimin Yi},
  \bibinfo{person}{Zelin Hu}, \bibinfo{person}{Bochao Liu},
  \bibinfo{person}{Peng Sun}, \bibinfo{person}{Haoyu Liu}, {et~al\mbox{.}}}
  \bibinfo{year}{2018}\natexlab{}.
\newblock \showarticletitle{Image matters: Visually modeling user behaviors
  using advanced model server}. In \bibinfo{booktitle}{\emph{Proceedings of the
  27th ACM International Conference on Information and Knowledge Management}}.
  \bibinfo{pages}{2087--2095}.
\newblock


\bibitem[\protect\citeauthoryear{Glowacka}{Glowacka}{2017}]%
        {glowacka2017bandit}
\bibfield{author}{\bibinfo{person}{Dorota Glowacka}.}
  \bibinfo{year}{2017}\natexlab{}.
\newblock \showarticletitle{Bandit Algorithms in Interactive Information
  Retrieval}. In \bibinfo{booktitle}{\emph{Proceedings of the {ACM} {SIGIR}
  International Conference on Theory of Information Retrieval, {ICTIR} 2017,
  Amsterdam, The Netherlands, October 1-4, 2017}},
  \bibfield{editor}{\bibinfo{person}{Jaap Kamps}, \bibinfo{person}{Evangelos
  Kanoulas}, \bibinfo{person}{Maarten de~Rijke}, \bibinfo{person}{Hui Fang},
  {and} \bibinfo{person}{Emine Yilmaz}} (Eds.). \bibinfo{publisher}{{ACM}},
  \bibinfo{pages}{327--328}.
\newblock
\urldef\tempurl%
\url{https://doi.org/10.1145/3121050.3121108}
\showDOI{\tempurl}


\bibitem[\protect\citeauthoryear{Glowacka}{Glowacka}{2019}]%
        {glowacka2019bandit}
\bibfield{author}{\bibinfo{person}{Dorota Glowacka}.}
  \bibinfo{year}{2019}\natexlab{}.
\newblock \showarticletitle{Bandit algorithms in recommender systems}. In
  \bibinfo{booktitle}{\emph{Proceedings of the 13th ACM Conference on
  Recommender Systems}}. \bibinfo{pages}{574--575}.
\newblock


\bibitem[\protect\citeauthoryear{He, Zhang, Ren, and Sun}{He
  et~al\mbox{.}}{2016}]%
        {he2016deep}
\bibfield{author}{\bibinfo{person}{Kaiming He}, \bibinfo{person}{Xiangyu
  Zhang}, \bibinfo{person}{Shaoqing Ren}, {and} \bibinfo{person}{Jian Sun}.}
  \bibinfo{year}{2016}\natexlab{}.
\newblock \showarticletitle{Deep residual learning for image recognition}. In
  \bibinfo{booktitle}{\emph{Proceedings of the IEEE conference on computer
  vision and pattern recognition}}. \bibinfo{pages}{770--778}.
\newblock


\bibitem[\protect\citeauthoryear{Kong, Shen, Lin, Mech, and Fowlkes}{Kong
  et~al\mbox{.}}{2016}]%
        {kong2016photo}
\bibfield{author}{\bibinfo{person}{Shu Kong}, \bibinfo{person}{Xiaohui Shen},
  \bibinfo{person}{Zhe~L. Lin}, \bibinfo{person}{Radom{\'{\i}}r Mech}, {and}
  \bibinfo{person}{Charless~C. Fowlkes}.} \bibinfo{year}{2016}\natexlab{}.
\newblock \showarticletitle{Photo Aesthetics Ranking Network with Attributes
  and Content Adaptation}. In \bibinfo{booktitle}{\emph{Computer Vision -
  {ECCV} 2016 - 14th European Conference, Amsterdam, The Netherlands, October
  11-14, 2016, Proceedings, Part {I}}} \emph{(\bibinfo{series}{Lecture Notes in
  Computer Science}, Vol.~\bibinfo{volume}{9905})},
  \bibfield{editor}{\bibinfo{person}{Bastian Leibe}, \bibinfo{person}{Jiri
  Matas}, \bibinfo{person}{Nicu Sebe}, {and} \bibinfo{person}{Max Welling}}
  (Eds.). \bibinfo{publisher}{Springer}, \bibinfo{pages}{662--679}.
\newblock
\urldef\tempurl%
\url{https://doi.org/10.1007/978-3-319-46448-0\_40}
\showDOI{\tempurl}


\bibitem[\protect\citeauthoryear{Li, Chu, Langford, and Schapire}{Li
  et~al\mbox{.}}{2010}]%
        {li2010contextual}
\bibfield{author}{\bibinfo{person}{Lihong Li}, \bibinfo{person}{Wei Chu},
  \bibinfo{person}{John Langford}, {and} \bibinfo{person}{Robert~E. Schapire}.}
  \bibinfo{year}{2010}\natexlab{}.
\newblock \showarticletitle{A contextual-bandit approach to personalized news
  article recommendation}. In \bibinfo{booktitle}{\emph{Proceedings of the 19th
  International Conference on World Wide Web, {WWW} 2010, Raleigh, North
  Carolina, USA, April 26-30, 2010}},
  \bibfield{editor}{\bibinfo{person}{Michael Rappa}, \bibinfo{person}{Paul
  Jones}, \bibinfo{person}{Juliana Freire}, {and} \bibinfo{person}{Soumen
  Chakrabarti}} (Eds.). \bibinfo{publisher}{{ACM}}, \bibinfo{pages}{661--670}.
\newblock
\urldef\tempurl%
\url{https://doi.org/10.1145/1772690.1772758}
\showDOI{\tempurl}


\bibitem[\protect\citeauthoryear{Liu, Lu, Yang, Zhao, Xu, Peng, Zhang, Niu,
  Zhu, Bao, et~al\mbox{.}}{Liu et~al\mbox{.}}{2020}]%
        {liu2020category}
\bibfield{author}{\bibinfo{person}{Hu Liu}, \bibinfo{person}{Jing Lu},
  \bibinfo{person}{Hao Yang}, \bibinfo{person}{Xiwei Zhao},
  \bibinfo{person}{Sulong Xu}, \bibinfo{person}{Hao Peng},
  \bibinfo{person}{Zehua Zhang}, \bibinfo{person}{Wenjie Niu},
  \bibinfo{person}{Xiaokun Zhu}, \bibinfo{person}{Yongjun Bao},
  {et~al\mbox{.}}} \bibinfo{year}{2020}\natexlab{}.
\newblock \showarticletitle{Category-Specific CNN for Visual-aware CTR
  Prediction at JD. com}. In \bibinfo{booktitle}{\emph{Proceedings of the 26th
  ACM SIGKDD International Conference on Knowledge Discovery \& Data Mining}}.
  \bibinfo{pages}{2686--2696}.
\newblock


\bibitem[\protect\citeauthoryear{Mo, Liu, Xiao, Li, and Jiang}{Mo
  et~al\mbox{.}}{2015}]%
        {mo2015image}
\bibfield{author}{\bibinfo{person}{Kaixiang Mo}, \bibinfo{person}{Bo Liu},
  \bibinfo{person}{Lei Xiao}, \bibinfo{person}{Yong Li}, {and}
  \bibinfo{person}{Jie Jiang}.} \bibinfo{year}{2015}\natexlab{}.
\newblock \showarticletitle{Image Feature Learning for Cold Start Problem in
  Display Advertising}. In \bibinfo{booktitle}{\emph{Proceedings of the
  Twenty-Fourth International Joint Conference on Artificial Intelligence,
  {IJCAI} 2015, Buenos Aires, Argentina, July 25-31, 2015}},
  \bibfield{editor}{\bibinfo{person}{Qiang Yang} {and}
  \bibinfo{person}{Michael~J. Wooldridge}} (Eds.). \bibinfo{publisher}{{AAAI}
  Press}, \bibinfo{pages}{3728--3734}.
\newblock
\urldef\tempurl%
\url{http://ijcai.org/Abstract/15/524}
\showURL{%
\tempurl}


\bibitem[\protect\citeauthoryear{Paszke, Gross, Chintala, Chanan, Yang, DeVito,
  Lin, Desmaison, Antiga, and Lerer}{Paszke et~al\mbox{.}}{2017}]%
        {paszke2017automatic}
\bibfield{author}{\bibinfo{person}{Adam Paszke}, \bibinfo{person}{Sam Gross},
  \bibinfo{person}{Soumith Chintala}, \bibinfo{person}{Gregory Chanan},
  \bibinfo{person}{Edward Yang}, \bibinfo{person}{Zachary DeVito},
  \bibinfo{person}{Zeming Lin}, \bibinfo{person}{Alban Desmaison},
  \bibinfo{person}{Luca Antiga}, {and} \bibinfo{person}{Adam Lerer}.}
  \bibinfo{year}{2017}\natexlab{}.
\newblock \showarticletitle{Automatic differentiation in pytorch}.
\newblock  (\bibinfo{year}{2017}).
\newblock


\bibitem[\protect\citeauthoryear{Precup}{Precup}{2000}]%
        {precup2000eligibility}
\bibfield{author}{\bibinfo{person}{Doina Precup}.}
  \bibinfo{year}{2000}\natexlab{}.
\newblock \showarticletitle{Eligibility traces for off-policy policy
  evaluation}.
\newblock \bibinfo{journal}{\emph{Computer Science Department Faculty
  Publication Series}} (\bibinfo{year}{2000}), \bibinfo{pages}{80}.
\newblock


\bibitem[\protect\citeauthoryear{Riquelme, Tucker, and Snoek}{Riquelme
  et~al\mbox{.}}{2018}]%
        {riquelme2018deep}
\bibfield{author}{\bibinfo{person}{Carlos Riquelme}, \bibinfo{person}{George
  Tucker}, {and} \bibinfo{person}{Jasper Snoek}.}
  \bibinfo{year}{2018}\natexlab{}.
\newblock \showarticletitle{Deep Bayesian Bandits Showdown: An Empirical
  Comparison of Bayesian Deep Networks for Thompson Sampling}. In
  \bibinfo{booktitle}{\emph{6th International Conference on Learning
  Representations, {ICLR} 2018, Vancouver, BC, Canada, April 30 - May 3, 2018,
  Conference Track Proceedings}}. \bibinfo{publisher}{OpenReview.net}.
\newblock
\urldef\tempurl%
\url{https://openreview.net/forum?id=SyYe6k-CW}
\showURL{%
\tempurl}


\bibitem[\protect\citeauthoryear{Russo and Roy}{Russo and Roy}{2014}]%
        {russo2014learning}
\bibfield{author}{\bibinfo{person}{Daniel Russo} {and}
  \bibinfo{person}{Benjamin~Van Roy}.} \bibinfo{year}{2014}\natexlab{}.
\newblock \showarticletitle{Learning to Optimize via Posterior Sampling}.
\newblock \bibinfo{journal}{\emph{Math. Oper. Res.}} \bibinfo{volume}{39},
  \bibinfo{number}{4} (\bibinfo{year}{2014}), \bibinfo{pages}{1221--1243}.
\newblock
\urldef\tempurl%
\url{https://doi.org/10.1287/moor.2014.0650}
\showDOI{\tempurl}


\bibitem[\protect\citeauthoryear{Schlimmer}{Schlimmer}{1981}]%
        {schlimmer1981mushroom}
\bibfield{author}{\bibinfo{person}{Jeff Schlimmer}.}
  \bibinfo{year}{1981}\natexlab{}.
\newblock \showarticletitle{Mushroom records drawn from the audubon society
  field guide to north american mushrooms}.
\newblock \bibinfo{journal}{\emph{GH Lincoff (Pres), New York}}
  (\bibinfo{year}{1981}).
\newblock


\bibitem[\protect\citeauthoryear{Schwartz, Bradlow, and Fader}{Schwartz
  et~al\mbox{.}}{2017}]%
        {schwartz2017customer}
\bibfield{author}{\bibinfo{person}{Eric~M Schwartz}, \bibinfo{person}{Eric~T
  Bradlow}, {and} \bibinfo{person}{Peter~S Fader}.}
  \bibinfo{year}{2017}\natexlab{}.
\newblock \showarticletitle{Customer acquisition via display advertising using
  multi-armed bandit experiments}.
\newblock \bibinfo{journal}{\emph{Marketing Science}} \bibinfo{volume}{36},
  \bibinfo{number}{4} (\bibinfo{year}{2017}), \bibinfo{pages}{500--522}.
\newblock


\bibitem[\protect\citeauthoryear{Schwarz, Wieschollek, and Lensch}{Schwarz
  et~al\mbox{.}}{2018}]%
        {schwarz2018will}
\bibfield{author}{\bibinfo{person}{Katharina Schwarz}, \bibinfo{person}{Patrick
  Wieschollek}, {and} \bibinfo{person}{Hendrik~PA Lensch}.}
  \bibinfo{year}{2018}\natexlab{}.
\newblock \showarticletitle{Will people like your image? learning the aesthetic
  space}. In \bibinfo{booktitle}{\emph{2018 IEEE Winter Conference on
  Applications of Computer Vision (WACV)}}. IEEE, \bibinfo{pages}{2048--2057}.
\newblock


\bibitem[\protect\citeauthoryear{Simonyan and Zisserman}{Simonyan and
  Zisserman}{2015}]%
        {simonyan2014very}
\bibfield{author}{\bibinfo{person}{Karen Simonyan} {and}
  \bibinfo{person}{Andrew Zisserman}.} \bibinfo{year}{2015}\natexlab{}.
\newblock \showarticletitle{Very Deep Convolutional Networks for Large-Scale
  Image Recognition}. In \bibinfo{booktitle}{\emph{3rd International Conference
  on Learning Representations, {ICLR} 2015, San Diego, CA, USA, May 7-9, 2015,
  Conference Track Proceedings}}, \bibfield{editor}{\bibinfo{person}{Yoshua
  Bengio} {and} \bibinfo{person}{Yann LeCun}} (Eds.).
\newblock
\urldef\tempurl%
\url{http://arxiv.org/abs/1409.1556}
\showURL{%
\tempurl}


\bibitem[\protect\citeauthoryear{Wang, Li, Cui, Zhang, and Mao}{Wang
  et~al\mbox{.}}{2011}]%
        {wang2011click}
\bibfield{author}{\bibinfo{person}{Xuerui Wang}, \bibinfo{person}{Wei Li},
  \bibinfo{person}{Ying Cui}, \bibinfo{person}{Ruofei Zhang}, {and}
  \bibinfo{person}{Jianchang Mao}.} \bibinfo{year}{2011}\natexlab{}.
\newblock \showarticletitle{Click-through rate estimation for rare events in
  online advertising}.
\newblock In \bibinfo{booktitle}{\emph{Online multimedia advertising:
  Techniques and technologies}}. \bibinfo{publisher}{IGI Global},
  \bibinfo{pages}{1--12}.
\newblock


\bibitem[\protect\citeauthoryear{Wang, Xu, Wu, Li, He, Hu, and Yan}{Wang
  et~al\mbox{.}}{2018}]%
        {wang2018telepath}
\bibfield{author}{\bibinfo{person}{Yu Wang}, \bibinfo{person}{Jixing Xu},
  \bibinfo{person}{Aohan Wu}, \bibinfo{person}{Mantian Li},
  \bibinfo{person}{Yang He}, \bibinfo{person}{Jinghe Hu}, {and}
  \bibinfo{person}{Weipeng~P Yan}.} \bibinfo{year}{2018}\natexlab{}.
\newblock \showarticletitle{Telepath: Understanding users from a human vision
  perspective in large-scale recommender systems}. In
  \bibinfo{booktitle}{\emph{Thirty-Second AAAI Conference on Artificial
  Intelligence}}.
\newblock


\bibitem[\protect\citeauthoryear{Yang, Li, Qin, and Ye}{Yang
  et~al\mbox{.}}{2020}]%
        {yang2020hierarchical}
\bibfield{author}{\bibinfo{person}{Mengyue Yang}, \bibinfo{person}{Qingyang
  Li}, \bibinfo{person}{Zhiwei Qin}, {and} \bibinfo{person}{Jieping Ye}.}
  \bibinfo{year}{2020}\natexlab{}.
\newblock \showarticletitle{Hierarchical Adaptive Contextual Bandits for
  Resource Constraint based Recommendation}. In
  \bibinfo{booktitle}{\emph{Proceedings of The Web Conference 2020}}.
  \bibinfo{pages}{292--302}.
\newblock


\bibitem[\protect\citeauthoryear{Yu, Zhang, He, Chen, Xiong, and Qin}{Yu
  et~al\mbox{.}}{2018}]%
        {yu2018aesthetic}
\bibfield{author}{\bibinfo{person}{Wenhui Yu}, \bibinfo{person}{Huidi Zhang},
  \bibinfo{person}{Xiangnan He}, \bibinfo{person}{Xu Chen}, \bibinfo{person}{Li
  Xiong}, {and} \bibinfo{person}{Zheng Qin}.} \bibinfo{year}{2018}\natexlab{}.
\newblock \showarticletitle{Aesthetic-based Clothing Recommendation}. In
  \bibinfo{booktitle}{\emph{Proceedings of the 2018 World Wide Web Conference
  on World Wide Web, {WWW} 2018, Lyon, France, April 23-27, 2018}},
  \bibfield{editor}{\bibinfo{person}{Pierre{-}Antoine Champin},
  \bibinfo{person}{Fabien~L. Gandon}, \bibinfo{person}{Mounia Lalmas}, {and}
  \bibinfo{person}{Panagiotis~G. Ipeirotis}} (Eds.).
  \bibinfo{publisher}{{ACM}}, \bibinfo{pages}{649--658}.
\newblock
\urldef\tempurl%
\url{https://doi.org/10.1145/3178876.3186146}
\showDOI{\tempurl}


\bibitem[\protect\citeauthoryear{Zhao, Li, Zhang, Wang, Jiang, Xu, Wang, and
  Ma}{Zhao et~al\mbox{.}}{2019}]%
        {zhao2019you}
\bibfield{author}{\bibinfo{person}{Zhichen Zhao}, \bibinfo{person}{Lei Li},
  \bibinfo{person}{Bowen Zhang}, \bibinfo{person}{Meng Wang},
  \bibinfo{person}{Yuning Jiang}, \bibinfo{person}{Li Xu},
  \bibinfo{person}{Fengkun Wang}, {and} \bibinfo{person}{Wei{-}Ying Ma}.}
  \bibinfo{year}{2019}\natexlab{}.
\newblock \showarticletitle{What You Look Matters?: Offline Evaluation of
  Advertising Creatives for Cold-start Problem}. In
  \bibinfo{booktitle}{\emph{Proceedings of the 28th {ACM} International
  Conference on Information and Knowledge Management, {CIKM} 2019, Beijing,
  China, November 3-7, 2019}}, \bibfield{editor}{\bibinfo{person}{Wenwu Zhu},
  \bibinfo{person}{Dacheng Tao}, \bibinfo{person}{Xueqi Cheng},
  \bibinfo{person}{Peng Cui}, \bibinfo{person}{Elke~A. Rundensteiner},
  \bibinfo{person}{David Carmel}, \bibinfo{person}{Qi~He}, {and}
  \bibinfo{person}{Jeffrey~Xu Yu}} (Eds.). \bibinfo{publisher}{{ACM}},
  \bibinfo{pages}{2605--2613}.
\newblock
\urldef\tempurl%
\url{https://doi.org/10.1145/3357384.3357813}
\showDOI{\tempurl}


\end{thebibliography}


\end{document}